\PassOptionsToPackage{dvipsnames,table}{xcolor} 
\documentclass[10pt,twocolumn,letterpaper]{article}

\usepackage[pagenumbers]{iccv} %

\makeatletter
\renewcommand{\maketag@@@}[1]{\hbox{\m@th\normalsize\normalfont#1}}%
\makeatother
\usepackage[ruled,noend,linesnumbered]{algorithm2e}
\usepackage{etoolbox}%
\providetoggle{beginFlag}
\settoggle{beginFlag}{true}

\SetAlgoSkip{SkipBeforeAndAfter}

\usepackage{makecell}
\usepackage{multirow}

\usepackage{bbding}

\definecolor{iccvblue}{rgb}{0.21,0.49,0.74}
\usepackage[pagebackref,breaklinks,colorlinks,allcolors=iccvblue]{hyperref}
\usepackage{array}     
\usepackage{graphicx}
\usepackage{subcaption}
\usepackage{float}
\usepackage{amsmath}
\usepackage{multirow}
\usepackage{colortbl}
\usepackage{fontawesome}

\usepackage{algpseudocode}
\usepackage{overpic}
\usepackage[utf8]{inputenc} 
\usepackage[T1]{fontenc}    
\usepackage{hyperref}       
\usepackage{url}            
\usepackage{booktabs}       
\usepackage{amsfonts}       
\usepackage{nicefrac}       
\usepackage{microtype}      
\usepackage{xcolor}         
\usepackage{amsmath}
\usepackage{amssymb}
\usepackage{enumitem}
\usepackage{makecell}
\usepackage{tikz}  
\usetikzlibrary{shapes.geometric, arrows.meta, calc} 
\usepackage{subcaption}
\usepackage{multirow}
\usepackage{multicol}
\usepackage{rotating}
\usepackage{comment}
\usepackage{wrapfig}
\usepackage{makecell}
\newtheorem{theorem}{Theorem}

\newcommand{\prompts}[1]{{\footnotesize\sffamily\itshape #1}} 

\title{Dive3D: Diverse Distillation-based Text-to-3D Generation \\ via Score Implicit Matching}

\author{Weimin Bai$^1$ \quad Yubo Li$^1$ \quad Wenzheng Chen$^1$ \quad Weijian Luo$^2$ \quad He Sun$^{1 *}$ \\
1. Peking University \quad
2. Xiaohongshu Inc \quad
$^*$Corresponding author
\\
\texttt{weiminbai@stu.pku.edu.cn; hesun@pku.edu.cn} \\  
    \vspace{0.25em}  
    Project page: \url{https://ai4scientificimaging.org/dive3d} 
}

\begin{document}

\twocolumn[{%
\renewcommand\twocolumn[1][]{#1}%
\vspace{-10mm}
\maketitle
\vspace{-10mm}
\begin{center}
    \centering
    \includegraphics[width=\linewidth]{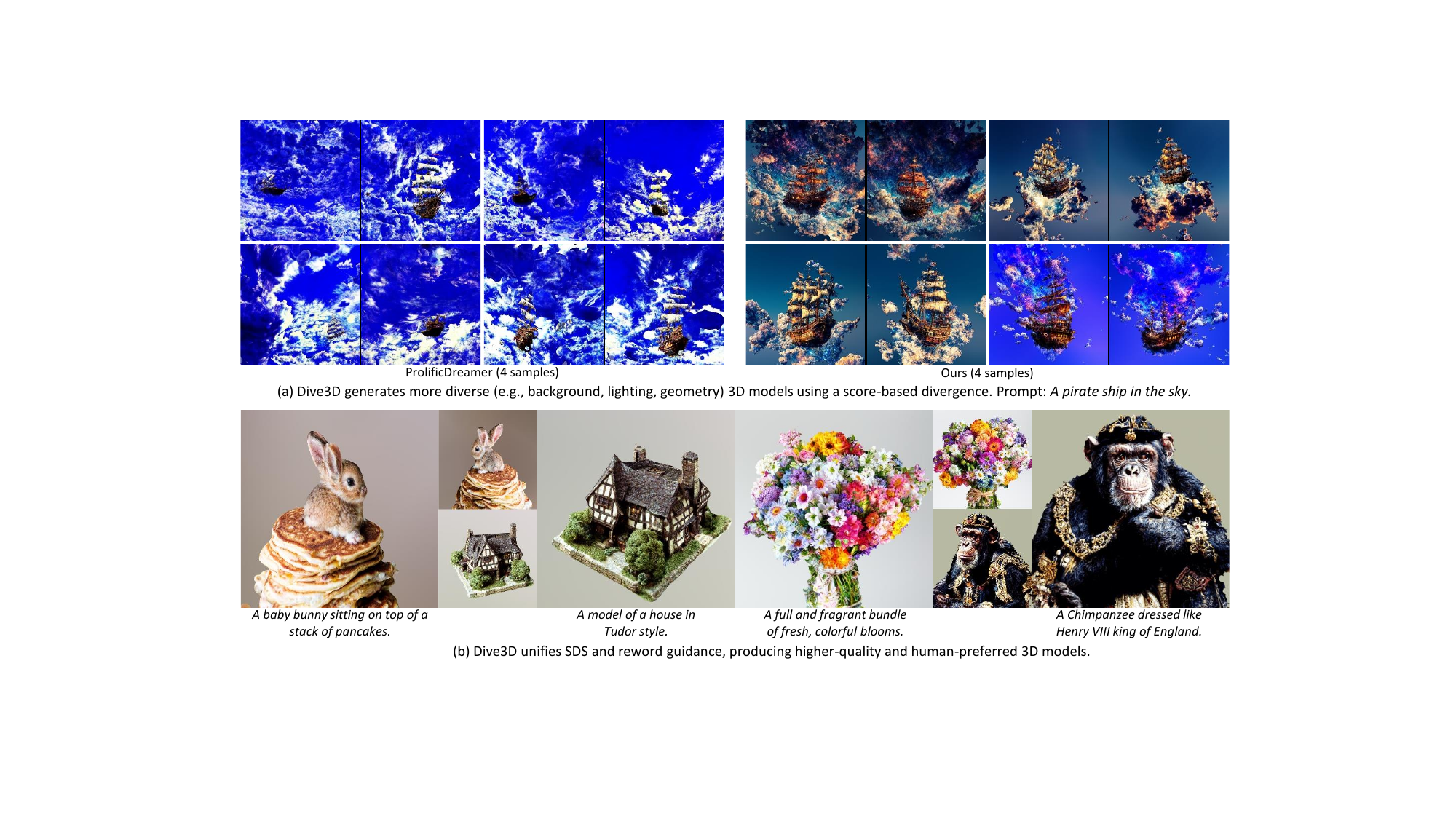}
    \vspace{-5mm}
    \captionof{figure}{
    We propose \textbf{Dive3D}, a novel text-to-3D generation framework that enables both \textbf{diverse} and \textbf{high-fidelity} 3D content creation.
\textbf{Top}: Compared to ProlificDreamer, which exhibits mode collapse under the same prompt, Dive3D generates a diverse set of realistic and semantically aligned 3D outputs.
\textbf{Bottom}: Dive3D unifies SDS and human-preference rewards within a divergence-based framework, effectively combining diffusion priors and reward signals to boost both visual quality and prompt alignment.
All results are represented as textured mesh assets. 
    }
\label{fig:teaser}
\end{center}%
}]


\begin{abstract}


Distilling pre-trained 2D diffusion models into 3D assets has driven remarkable advances in text-to-3D synthesis. 
However, existing methods typically rely on Score Distillation Sampling (SDS) loss, which involves asymmetric KL divergence—a formulation that inherently favors mode-seeking behavior and limits generation diversity. 
In this paper, we introduce Dive3D, a novel text-to-3D generation framework that replaces KL-based objectives with Score Implicit Matching (SIM) loss, a score-based objective that effectively mitigates mode collapse. 
Furthermore, Dive3D integrates both diffusion distillation and reward-guided optimization under a unified divergence perspective. Such reformulation, together with SIM loss, yields significantly more diverse 3D outputs while improving text alignment, human preference, and overall visual fidelity.
%
%
We validate Dive3D across various 2D-to-3D prompts and find that it consistently outperforms prior methods in qualitative assessments, including diversity, photorealism, and aesthetic appeal. We further evaluate its performance on the GPTEval3D benchmark, comparing against nine state-of-the-art baselines. Dive3D also achieves strong results on quantitative metrics, including text–asset alignment, 3D plausibility, text–geometry consistency, texture quality, and geometric detail.


\end{abstract}    
\section{Introduction}
\label{sec:introduction}

Text-to-3D generation—the task of creating 3D contents from natural language descriptions—has attracted enormous interest~\cite{poole2022dreamfusion,shi2023mvdream,lin2023magic3d}, due to its broad applications in vision and graphics.
Recent advances, such as 3D representations\citep{mildenhall2021nerf, kerbl20233d}, large-scale pre-trained vision-language models\citep{radford2021learning}, advanced text-to-image diffusion and flow models\citep{rombach2022high}, and differentiable rendering techniques, have further accelerated progress in this field. 
In particular, powerful text-to-image diffusion models such as Stable Diffusion series\citep{rombach2022high, ramesh2022hierarchical, saharia2022photorealistic}, lay a strong foundation for text-driven 3D synthesis: by leveraging pre-trained 2D diffusion priors and multi-view rendering, one can optimize a 3D asset so that its renderings align with a given text prompt. This capability opens new avenues for 3D content creation, 
enabling even non-experts to ``describe and create’’ novel 3D assets in freestyles.

Several paradigms have emerged to tackle text-to-3D generation. Diffusion distillation-based methods—exemplified by Score Distillation Sampling (SDS) in DreamFusion~\cite{poole2022dreamfusion}—optimize 3D representations by aligning multi-view renderings with pre-trained text-to-image diffusion priors~\cite{rombach2022high}. Reward-guided approaches~\cite{mohammad2022clip, jain2022zero} further refine these approaches by directly incorporating human-preference or CLIP-based rewards, boosting both semantic alignment and perceived quality. 
Despite their impressive fidelity and text alignment, both diffusion-distillation and reward-guided methods suffer from a critical limitation: \emph{limited generative diversity}. Even when prompted with intentionally vague or open-ended descriptions, current models tend to converge on a narrow set of similar outputs. 


We analyze this limitation and trace its root to the utilization of Kullback–Leibler (KL) divergence. Specifically, the objectives optimized by both SDS and reward-based methods can be reformulated as minimizing an asymmetric KL divergence, which shares a fundamental limitation: KL divergence inherently encourages mode-seeking behavior by penalizing samples that deviate from high-density regions of the target distribution. As a result, the generative model tends to collapse to a few dominant modes, severely suppressing output diversity.


In this paper, we present Dive3D, a novel framework that replaces KL-based objectives with Score Implicit Matching (SIM)—a score-based divergence loss that directly matches the gradient fields of the probability density of generated contents and the diffusion prior. This formulation avoids the mode-seeking tendencies of KL and encourages exploration of multiple high-probability regions, thereby promoting diversity without sacrificing fidelity or alignment. 
Furthermore, Dive3D unifies both diffusion distillation and reward-guided optimization under a divergence-based perspective. Combined with SIM loss, this formulation enables a principled integration of diffusion priors, human preferences, and diversity-promoting objectives within a single framework. As a result, Dive3D generates 3D assets that are not only realistic and well-aligned with text prompts, but also significantly more diverse.

Through extensive experiments on standard text-to-3D benchmarks, we demonstrate that Dive3D achieves state-of-the-art performance, substantially outperforming existing SDS- and reward-based approaches across visual fidelity, prompt adherence, and generative diversity.

\section{Related Works}
\label{sec:related}

\paragraph{Diffusion Distillation Based Methods}
Diffusion distillation methods~\citep{luo2023comprehensive} leverage pre-trained text-to-image diffusion models~\citep{rombach2022high, saharia2022photorealistic, balaji2022ediff} to guide the optimization of 3D representations by aligning rendered views with diffusion priors. This line of work was pioneered by Score Distillation Sampling (SDS) in DreamFusion~\citep{poole2022dreamfusion}, which initiated an era of 3D synthesis by transferring the knowledge embedded in 2D diffusion priors. However, these diffusion-driven optimization techniques typically rely on minimizing KL divergences~\citep{poole2022dreamfusion,wang2024prolificdreamer,luo2024diff}, often resulting in mode-seeking behavior where generated 3D objects collapse into a single plausible solution with limited diversity. Moreover, the straightforward use of 2D diffusion priors can introduce visual artifacts such as over-saturated colors, overly smooth geometry, and even Janus artifacts~\citep{jain2022zero,chen2023fantasia3d,wang2024prolificdreamer}. To address these challenges, recent studies have explored various improvements, including timestep annealing~\citep{huang2023dreamtime,wang2024prolificdreamer,zhu2023hifa}, coarse-to-fine training~\citep{lin2023magic3d, wang2024prolificdreamer, chen2023fantasia3d}, component analysis~\citep{katzir2023noise}, and formulation refinements~\citep{zhu2023hifa, wang2024prolificdreamer, liang2023luciddreamer, tang2023stable, wang2023taming, yu2023text, armandpour2023re, wu2024consistent3d, yan2024fsd}. Additional efforts have focused on geometry-texture disentanglement~\citep{chen2023fantasia3d, ma2023geodream, wang2024prolificdreamer} and mitigating the multi-face (Janus) problem by replacing text-to-image diffusion with novel view synthesis or multi-view diffusion~\citep{liu2023zero, long2023wonder3d, liu2023syncdreamer, weng2023consistent123, ye2023consistent, wang2023imagedream, shi2023mvdream}. Notably, diffusion distillation has also seen rapid progress in other domains, such as one-step diffusion models~\citep{luo2023diff,luo2024one,yin2024one,zhou2024score,geng2024consistency,huang2024flow} and various related approaches~\citep{nie2022diffusion,zhang2023enhancing,zhang2023purify++,luo2023entropy}.

\paragraph{Reward Optimization based Methods.}
Another category of approaches optimizes 3D outputs directly using reward models, such as visual-language alignment losses or human-preference reward models instead of (or in addition to) a diffusion prior. Early methods like CLIP-Mesh~\citep{mohammad2022clip} and DreamFields~\citep{jain2022zero}  directly maximize the CLIP score~\citep{radford2021learning} between rendered images and the text prompt, enabling zero-shot text-to-3D without 3D datasets. While conceptually simple, these CLIP-guided approaches often yielded suboptimal geometry or texture (e.g. unrealistic shapes) and required expensive optimization. More recently, DreamReward~\citep{ye2024dreamreward}  uses a learned internal 3D preference-reward model (Reward3D) trained on internally collected human feedback data to guide generation. DreamReward improves alignment of generated shapes with user intent, achieving better text relevance as judged by the reward function. Reward-based methods explicitly push for semantic or aesthetic alignment, but relying solely on them can compromise visual fidelity if the reward is not perfectly aligned with 3D realism (e.g. CLIP might encourage implausible textures). They may also require costly human data collection to train the internal 3D reward model.

\paragraph{Feed-forward Methods.}
Feed-forward methods train neural networks to directly generate 3D content from text using large synthetic 3D datasets or cross-modal supervision. For example, CLIP-Forge~\citep{sanghi2022clip} and CLIP-Sculptor~\citep{sanghi2023clip} leverage CLIP embeddings for zero-shot text-to-shape generation. More recently, advances in large reconstruction models (LRMs)\citep{hong2023lrm} have enabled rapid 3D model prediction from single or sparse-view images, inspiring these developments of methods like Instant3D\citep{li2023instant3d} and Turbo3D~\citep{hu2024turbo3d} that first generate multi-view images from text and then use a feed-forward 3D reconstructor (trained on synthetic data) to instantly produce representations such as NeRF or 3D Gaussian Splatting. However, the quality of these approaches depends heavily on the underlying text-to-multi-view generator, often recasting the challenge as one of diffusion distillation or reward-based optimization.
\section{Preliminary}
\label{sec:preliminary}

In this section, we review the key concepts and mathematical formulations underlying our work. We first describe text-to-image diffusion models, then explain how these models are adapted for text-to-3D generation via diffusion distillation, and finally review reward-guided text-to-3D methods.

\subsection{Text-to-Image Diffusion Models}

Diffusion models~\cite{sohl2015deep, ho2020denoising, song2020score} are a class of generative models that iteratively transform noise into data using a stochastic process. Let $\boldsymbol{x}_0 \sim q_{\text{data}}(\boldsymbol{x})$ denote a data sample. The forward diffusion process corrupts $\boldsymbol{x}_0$ by gradually adding noise described by the stochastic differential equation (SDE):
\begin{equation}
    d\boldsymbol{x}_t = \boldsymbol{F}(\boldsymbol{x}_t, t)\, dt + G(t)\, d\boldsymbol{w}_t,\quad t \in [0, T],
\end{equation}
where $\boldsymbol{F}(\boldsymbol{x}_t, t)$ is a drift function, $G(t)$ is a scalar-valued diffusion coefficient, and $\boldsymbol{w}_t$ denotes a standard Wiener process.
To generate samples, the reverse diffusion process is used to progressively denoise an initial noise sample \citep{song2020score,song2020denoising,zhang2022fast,liu2022pseudo,lu2022dpm,xue2023sa}. 

The marginal core function $\nabla_{\boldsymbol{x}_t} \log p_t(\boldsymbol{x}_t)$ is typically approximated by a continuous-indexed neural network $s_\phi(\boldsymbol{x}_t, t)$. This score network is trained using the weighted denoising score matching objective:
\begin{equation}
    \mathcal{L}(\phi)= \mathbb{E}_{t,\boldsymbol{x}_0,\epsilon}\left[ \lambda(t) \left\| s_\phi\Big(\alpha_t\boldsymbol{x}_0 + \sigma_t \epsilon, t\Big) + \frac{\epsilon}{\sigma_t} \right\|_2^2 \right],
    \label{eq:score_matching}
\end{equation}
where $\epsilon \sim \mathcal{N}(0, \mathbf{I})$, and the functions $\alpha_t$ and $\sigma_t$ are determined by the noise schedule. 

By conditioning on text inputs, these diffusion models can be extended to text-to-image synthesis. In this setting, a conditional score network $s_\phi(\boldsymbol{x}_t, y, t) \approx \nabla_{\boldsymbol{x}_t} \log p_t(\boldsymbol{x}_t|y)$ is used, where $y$ is the text prompt describing the image content. Popular models such as Stable Diffusion~\citep{rombach2022high} and MVDiffusion~\citep{shi2023mvdream} have demonstrated that this approach yields high-quality, semantically aligned images.

\subsection{Text-to-3D Generation by Diffusion Distillation}

A prevalent paradigm for text-to-3D synthesis leverages pretrained text-to-image diffusion models to guide the optimization of a 3D representation. Let $g(\theta, c)$ be a differentiable renderer that maps the 3D parameters $\theta$ to a 2D image under camera pose $c$, $q_\theta(\boldsymbol{x}_t|c)$ be the distribution of rendered images at diffusion time $t$, and $p(\boldsymbol{x}_t|y^c)$ be the target conditional distribution given a view-dependent text prompt $y^c$ defined by a pretrained diffusion model. The loss that aligns each rendered view of the 3D model with the conditional diffusion prior can be formulated as:
\begin{equation}\label{eqn:sds_loss_1}
    \mathcal{L}_{\mathrm{CDP}}(\theta) = \mathbb{E}_{t,c}\left[ \omega(t)\, D_{\mathrm{KL}}\Big( q_\theta(\boldsymbol{x}_t|c) \,\big\|\, p(\boldsymbol{x}_t|y^c) \Big) \right],
\end{equation}
where $\omega(t)$ is a weighting function. In practice, the gradient of loss \eqref{eqn:sds_loss_1} writes (please refer to \citet{luo2024diff} and \citet{wang2023score} for a comprehensive derivation):
\begin{equation} \label{eq:sds_grad}
    \nabla_\theta \mathcal{L}_{\mathrm{CDP}}(\theta) \approx \mathbb{E}_{t,\epsilon,c}\left[ \omega(t) \left( \epsilon_\phi(\boldsymbol{x}_t, y^c, t) - \epsilon \right) \frac{\partial g(\theta,c)}{\partial \theta} \right],
\end{equation}
where $\epsilon_\phi(\boldsymbol{x}_t, y^c, t)=-\sigma_t s_\phi(\boldsymbol{x}_t, y^c, t)$ is the noise prediction of the diffusion model.

The Score Distillation Sampling (SDS) loss, introduced in DreamFusion~\cite{poole2022dreamfusion}, improves generation quality by employing classifier-free guidance (CFG)\citep{ho2022classifier, ahn2024self, karras2024guiding, li2024self}, which replaces the original conditional score in Eq.\ref{eq:sds_grad} with a weighted difference between the conditional and unconditional score estimates,
\begin{equation} \label{eq:VDS}
\begin{split}
    &\hat{\epsilon}_\phi(\boldsymbol{x}_t,y^c,t) = (1+\gamma)\epsilon_\phi(\boldsymbol{x}_t,y^c,t) - \gamma\, \epsilon_\phi(\boldsymbol{x}_t,t)\\
    &= \epsilon_\phi(\boldsymbol{x}_t,y^c,t) + \gamma \Big(\epsilon_\phi(\boldsymbol{x}_t,y^c,t) - \epsilon_\phi(\boldsymbol{x}_t,t)\Big)\\
    &= -\sigma_t\Big[s_\phi(\boldsymbol{x}_t,y^c,t) + \gamma \Big(s_\phi(\boldsymbol{x}_t,y^c,t) - s_\phi(\boldsymbol{x}_t,t)\Big)\Big],
\end{split}
\vspace{-1em}
\end{equation}
\begin{equation} \label{eq:sds_grad2}
    \nabla_\theta \mathcal{L}_{\mathrm{SDS}}(\theta) \approx \mathbb{E}_{t,\epsilon,c}\left[ \omega(t) \left( \hat{\epsilon}_\phi(\boldsymbol{x}_t, y^c, t) - \epsilon \right) \frac{\partial g(\theta,c)}{\partial \theta} \right].
\end{equation}
This adjustment is equivalent to incorporating an additional regularization term into the Score Distillation Sampling (SDS) loss - the so-called CFG-reward as introduced by \citep{luo2024diffstar} ($\mathcal{L}_{\mathrm{SDS}}=\mathcal{L}_{\mathrm{CDP}}+\gamma \mathcal{L}_{\mathrm{CFR}}$), effectively acting as an implicit likelihood term that better aligns the generated image with the text prompt and enforces pose constraints. Increasing the weighting factor $\gamma$ strengthens this alignment, thereby improving the semantic calibration of the 3D renderings.


\label{sec:method}
\begin{figure*}[t]
    \centering
    \includegraphics[width=1\textwidth]{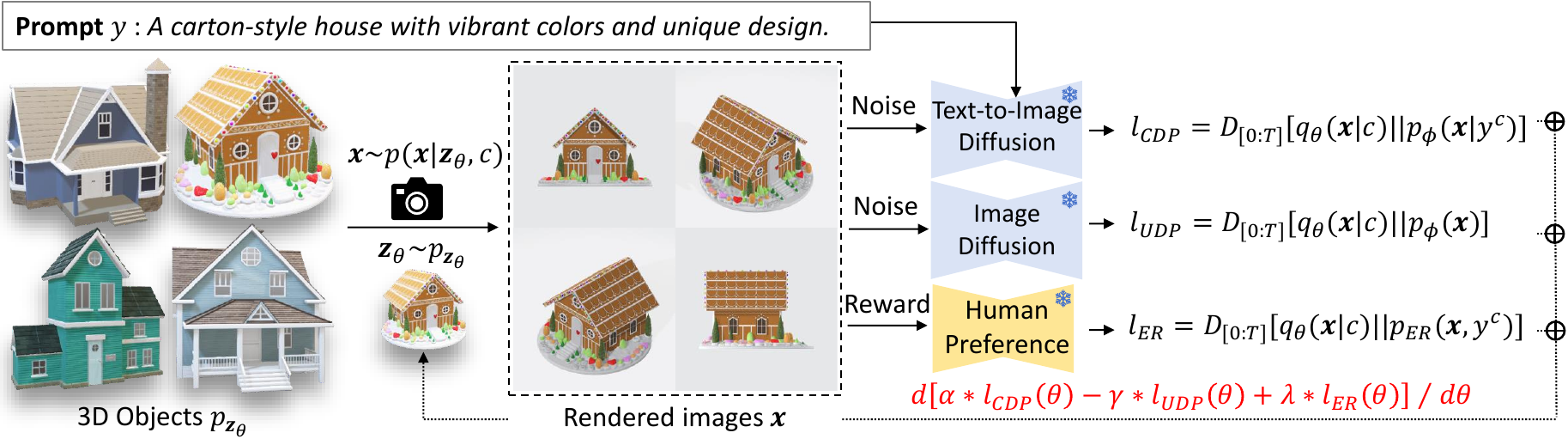}
    \caption{\textbf{Overview of Dive3D.} 
    {
    {Dive3D} reformulates both SDS loss and human-preference rewards within a unified divergence-based framework, revealing that both rely on KL divergence, $D_{\mathrm{KL}}(\cdot, \cdot)$—an asymmetric objective that inherently favors mode-seeking and leads to model collapse and limited diversity. 
    Dive3D replaces this with a \textbf{Score Implicit Matching}-based divergence, $D_{[0,T]}(\cdot, \cdot)$, which aligns score fields rather than probability densities, effectively mitigating mode collapse and enabling more diverse and faithful 3D generation.
    }}
    \label{fig:overview}
    \vspace{-1em}
\end{figure*}

\subsection{Text-to-3D Generation via Reward Guidance}

An alternative approach leverages reward signals to steer the generation of 3D content. Pioneering works such as DreamFields~\cite{jain2022zero}, CLIP-Mesh~\cite{mohammad2022clip}, and X-Mesh~\cite{ma2023x} leverage CLIP scores~\cite{radford2021learning} to align 3D representations with text prompts. In these methods, a reward function is defined as:
\begin{equation}
    r(y, x, c) = f\big(g(\theta, c)\big)^\top h(y^c),
\end{equation}
where $f(\cdot)$ and $h(\cdot)$ are embedding functions for images and text, respectively, and $g(\theta,c)$ is the rendered image. Maximizing this reward encourages the 3D model to generate outputs that are semantically aligned with the text. 

Recent methods, such as DreamReward\cite{ye2024dreamreward}, combine the SDS loss with reward-based signals to further enhance semantic alignment and human-preference consistency. For example, DreamReward modifies the SDS loss as:
\begin{equation} \label{eq:RreamReward}
    \mathcal{L}_{\mathrm{Reward}}(\theta) = \mathcal{L}_{\mathrm{SDS}}(\theta) - \lambda \, 
    \mathbb{E}_{t,c,\, \boldsymbol{x}_t}\Big[ \omega(t)\, r\big(y^c, \hat{x}_0(\boldsymbol{x}_t)\big) \Big],
\end{equation}
where $\boldsymbol{x}_t\sim q_\theta(\boldsymbol{x}_t|c)$, $\hat{x}_0 = \frac{1}{\alpha_t}\left[\boldsymbol{x}_t - \sigma_t\epsilon_\phi(\boldsymbol{x}_t, y, t)\right]$ is an estimate of the denoised image, and $\lambda$ balances the influence of the reward. Similar to Eq.~\ref{eq:VDS}, the reward function acts as an additional regularization term in SDS-based 3D generation. 

\section{Method}

In this section, we introduce \emph{Dive3D}, a principled framework that boosts both diversity and fidelity in text-to-3D synthesis by replacing KL-divergence guidance with score-based divergence optimization (see Fig.\ref{fig:overview}). In Sec.\ref{subsec:unify}, we demonstrate that existing SDS and reward losses are both linear combinations of KL divergences—and thus prone to mode collapse and mode-seeking. Then, in Sec.~\ref{subsec:score-divergence}, we present our score-based divergence formulation, which overcomes these limitations and delivers significantly more varied and higher-quality 3D outputs.

\subsection{SDS and Reward Guidance are Both KL Divergences} \label{subsec:unify}
\paragraph{\textbf{The SDS Loss.}}  
The classifier-free guidance in the  SDS loss (Eqs.~\ref{eq:VDS}–\ref{eq:sds_grad2}) can be rewritten as
\begin{equation} \label{eq:cfr1}
s_\phi(\boldsymbol{x}_t, y, t) - s_\phi(\boldsymbol{x}_t, t) \approx \nabla_{\boldsymbol{x}_t}\log p(\boldsymbol{x}_t|y) - \nabla_{\boldsymbol{x}_t}\log p(\boldsymbol{x}_t).
\end{equation}
Substituting Eq.~\ref{eq:cfr1} into Eq.~\ref{eq:sds_grad2} and integrating, the SDS loss can be expressed as the difference between two KL divergence terms:
\begin{equation}\label{eqn:cfg_loss_alternative}
\begin{split}
\mathcal{L}_{\mathrm{SDS}}(\theta) &= (1+\gamma)\,\mathcal{L}_{\mathrm{CDP}}(\theta) - \gamma\, \mathcal{L}_{\mathrm{UDP}}(\theta) \\
&= (1+\gamma)\,\mathbb{E}_{t,c}\Bigg[ \omega(t)\, D_{\mathrm{KL}}\Big( q_\theta(\boldsymbol{x}_t|c) \,\big\|\, p(\boldsymbol{x}_t|y^c) \Big) \Bigg] \\
&\quad - \gamma\, \mathbb{E}_{t,c}\Bigg[ \omega(t)\, D_{\mathrm{KL}}\Big( q_\theta(\boldsymbol{x}_t|c) \,\big\|\, p(\boldsymbol{x}_t) \Big) \Bigg].
\end{split}
\end{equation}
This formulation makes explicit that the SDS loss balances two KL divergences—one that promotes prompt fidelity (\(\mathcal{L}_{\mathrm{CDP}}\)) and one that modulates diversity via the unconditional prior (\(\mathcal{L}_{\mathrm{UDP}}\)). Increasing $\gamma$ strengthens text–image alignment but narrows diversity by shrinking the effective entropy.

\paragraph{\textbf{The Explicit Reward Loss.}}  

Assuming the reward defines an exponential distribution,
\begin{equation}
p_{\mathrm{ER}}(y^c,x_t) \propto \exp\Big(r\big(y^c, \hat{x}_0(\boldsymbol{x}_t)\big)\Big),
\end{equation}
the explicit reward loss in Eq.~\ref{eq:RreamReward} can likewise be interpreted as a KL divergence. 
\vspace{-0.5em}
\begin{equation}
\begin{split}
\mathcal{L}_{\mathrm{ER}}(\theta) &= \mathbb{E}_{t,c}\Big[ \omega(t)\, D_{\mathrm{KL}}\Big( q_\theta(\boldsymbol{x}_t|c) \,\big\|\, p_{\mathrm{ER}}(y^c,\boldsymbol{x}_t) \Big) \Big] \\
&= \mathbb{E}_{t,c,\, \boldsymbol{x}_t}\Big[ \omega(t)\,\Big(\log q_\theta(\boldsymbol{x}_t|c) - \log p_{\mathrm{ER}}(y^c,\boldsymbol{x}_t) \Big) \Big] \\
&= \text{constant} - \mathbb{E}_{t,c,\, \boldsymbol{x}_t}\Big[ \omega(t)\, r\big(y^c, \hat{x}_0(\boldsymbol{x}_t)\big) \Big],
\end{split}
\end{equation}
where the first term is a constant because the distribution \(q_\theta(x_t|c)\) is typically a uniformly-distributed collection of \(N\) particles (i.e., \(q_\theta(x_t|c)=1/N\)). Serving as a measure of the joint distribution of prompts and images, the explicit reward loss not only enhances text alignment during 3D generation but also provides the flexibility to incorporate additional criteria, such as human preference~\cite{xu2023imagereward, murray2012ava}, photorealism~\cite{kirstain2023pick}, and geometric consistency~\cite{ye2024dreamreward}.

\paragraph{\textbf{Unified KL Divergence Framework.}}
Collecting these components, we can unify all loss terms in the diffusion- or reward-based text-to-3D generation framework by defining three core KL-based terms:
\begin{equation} \label{eq:dive3D_loss1}
\begin{split}
\mathcal{L}_{\mathrm{CDP}}(\theta) &= \mathbb{E}_{t,c}\Big[ \omega(t) D_{\mathrm{KL}}\Big( q_\theta(\boldsymbol{x}_t|c) \,\big\|\, p(\boldsymbol{x}_t|y^c) \Big) \Big],\\ 
\mathcal{L}_{\mathrm{UDP}}(\theta) &= \mathbb{E}_{t,c}\Big[ \omega(t) D_{\mathrm{KL}}\Big( q_\theta(\boldsymbol{x}_t|c) \,\big\|\, p(\boldsymbol{x}_t) \Big) \Big],\\
\mathcal{L}_{\mathrm{ER}}(\theta) &= \mathbb{E}_{t,c}\Big[ \omega(t) D_{\mathrm{KL}}\Big( q_\theta(\boldsymbol{x}_t|c) \,\big\|\, p_{\mathrm{ER}}(y^c, \boldsymbol{x}_t) \Big) \Big]. \\
\end{split}
\end{equation} 
Both SDS and reward-guided objectives are simply linear combinations of these divergences:
\begin{equation} \label{eq:dive3D_loss2}
\begin{split}
\mathcal{L}_{\mathrm{SDS}} &= (1+\gamma) \mathcal{L}_{\mathrm{CDP}} - \gamma \mathcal{L}_{\mathrm{UDP}},\\
\mathcal{L}_{\mathrm{Reward}} &=  \mathcal{L}_{\mathrm{SDS}} + \lambda \mathcal{L}_{\mathrm{ER}}.
\end{split}
\end{equation} 

This unified view permits flexible tuning of the weights on each term (see Appendix), yielding higher‐fidelity generations. However, both theory and experiments~\cite{luo2025one, zhou2024score, zhou2024long} show that relying on the inherently asymmetric KL divergence ($D_{\mathrm{KL}}(q|p)\neq D_{\mathrm{KL}}(p|q)$) destabilizes training and induces mode‐seeking, thereby constraining the diversity of generated 3D assets.

\begin{figure*}[!ht]
    \centering
    \setlength{\tabcolsep}{1pt}
    \setlength{\fboxrule}{1pt}
    \resizebox{0.99\textwidth}{!}{
    \begin{tabular}{c}
    \begin{tabular}{cc|cc|cc|cc}
        \multicolumn{2}{c}{{DreamFusion~\cite{poole2022dreamfusion}}} &
        \multicolumn{2}{c}{{Fantasia3D~\cite{chen2023fantasia3d}}} &
        \multicolumn{2}{c}{{ProlificDreamer~\cite{wang2024prolificdreamer}}} &
        \multicolumn{2}{c}{\textbf{Dive3D (Ours)}}
        \\
        \includegraphics[width=0.15\textwidth]{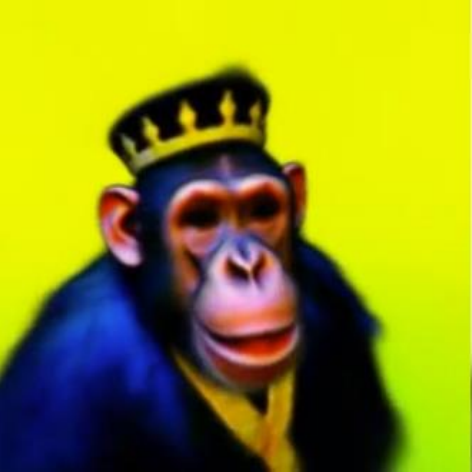} &
        \includegraphics[width=0.15\textwidth]{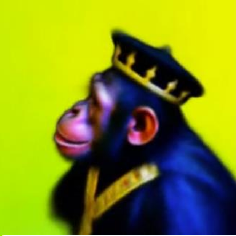} &
        \includegraphics[width=0.15\textwidth]{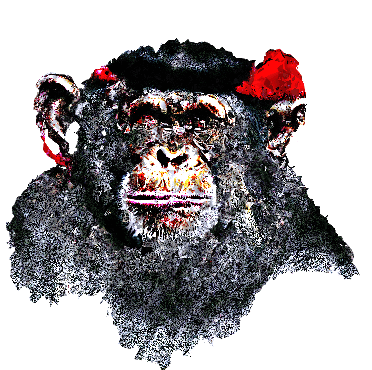} &
        \includegraphics[width=0.15\textwidth]{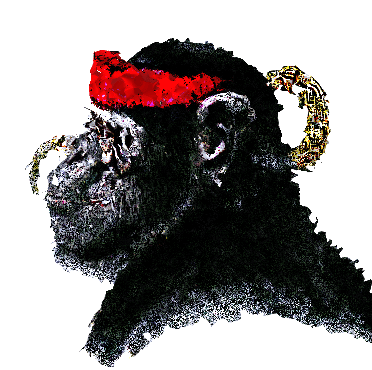} &
        \includegraphics[width=0.15\textwidth]{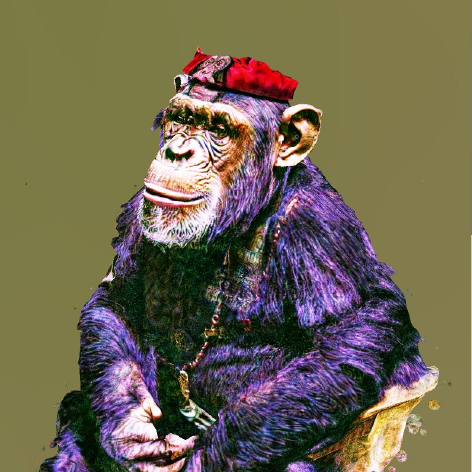} &
        \includegraphics[width=0.15\textwidth]{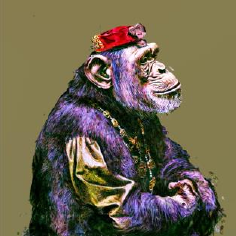} &
        \includegraphics[width=0.15\textwidth]{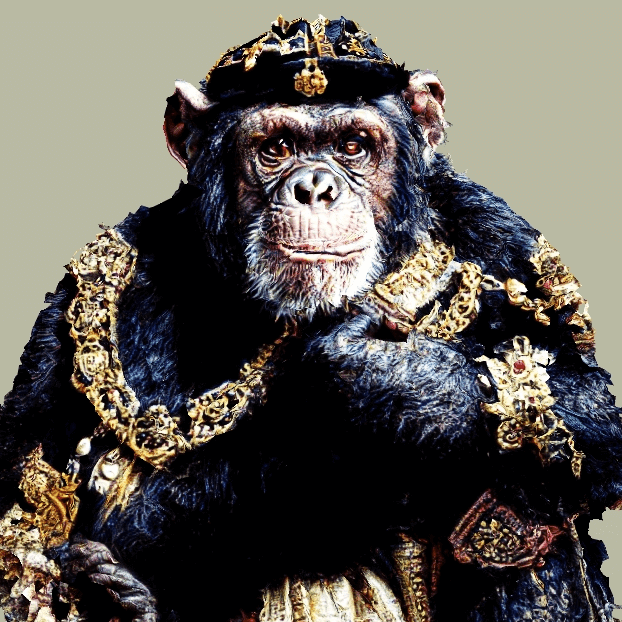} &
        \includegraphics[width=0.15\textwidth]{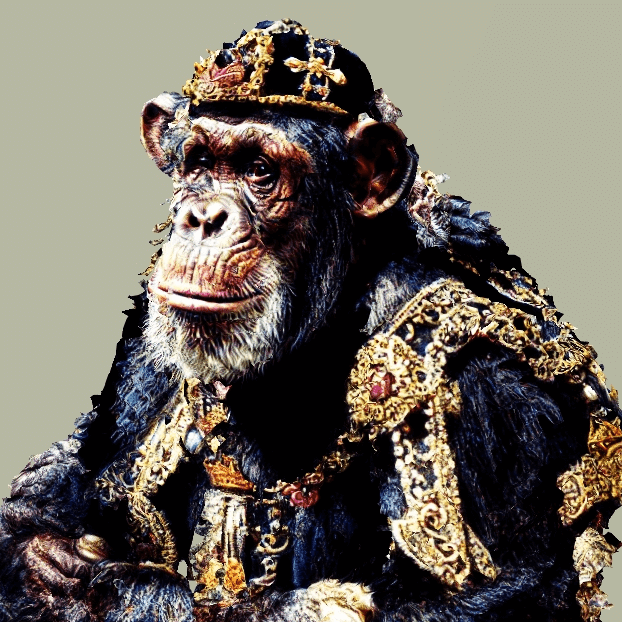} 
        \vspace{-0.5em}
        \\
        \multicolumn{8}{c}{{\prompts{A chimpanzee dressed like \textcolor{red}{Henry VIII king of England}.}}}
        \\
        \includegraphics[width=0.15\textwidth]{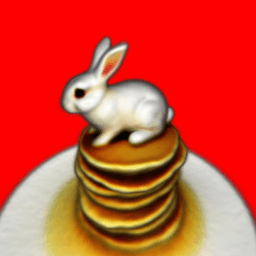} &
        \includegraphics[width=0.15\textwidth]{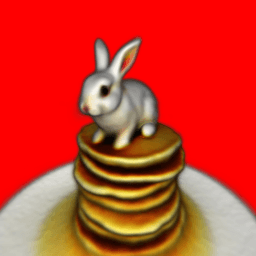} &
        \includegraphics[width=0.15\textwidth]{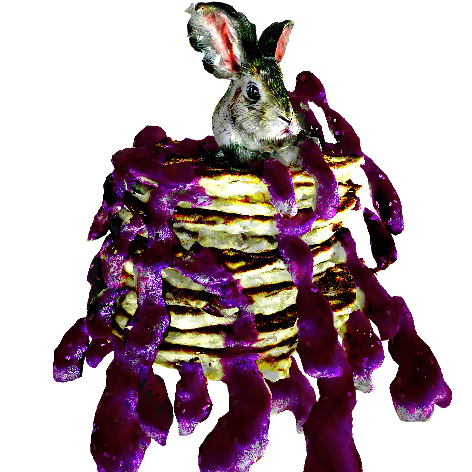} &
        \includegraphics[width=0.15\textwidth]{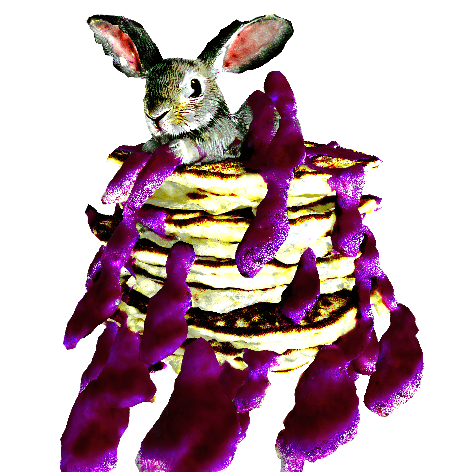} &
        \includegraphics[width=0.15\textwidth]{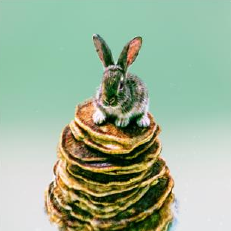} &
        \includegraphics[width=0.15\textwidth]{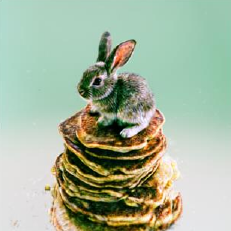} &
        \includegraphics[width=0.15\textwidth]{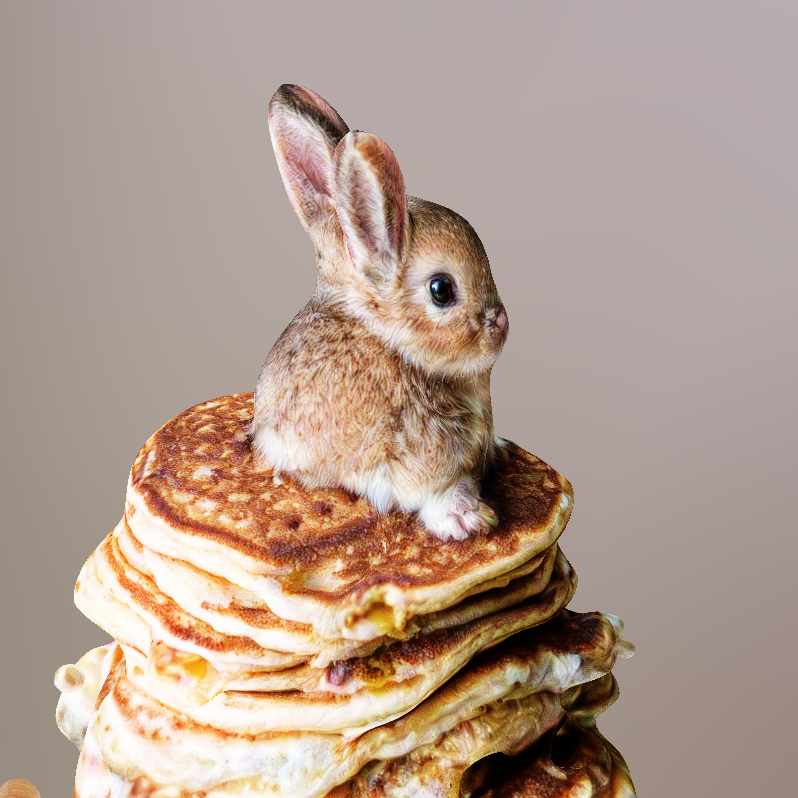} &
        \includegraphics[width=0.15\textwidth]{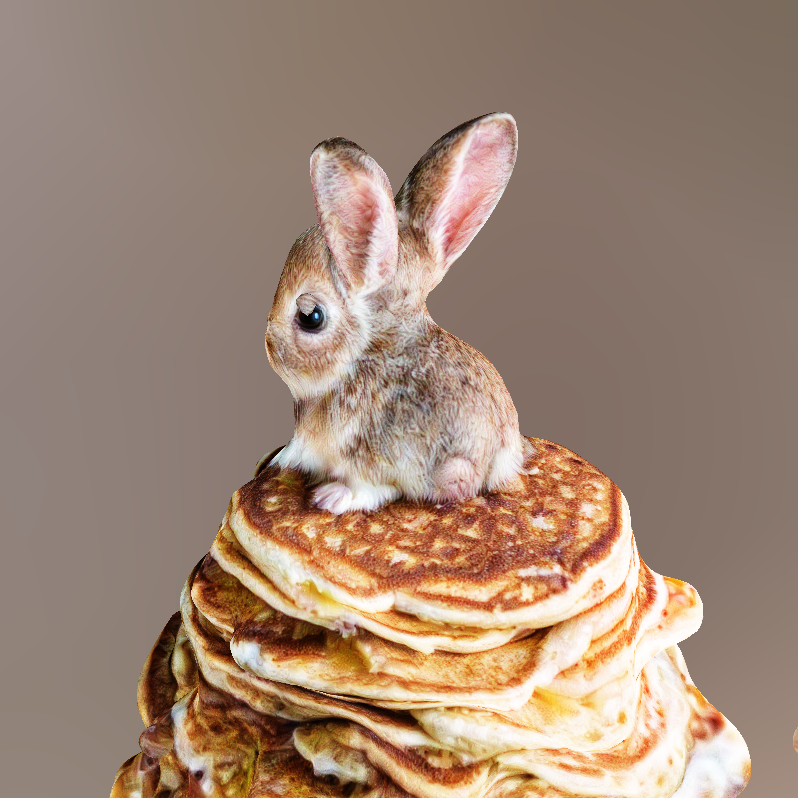} 
        \vspace{-0.5em}
        \\
        \multicolumn{8}{c}{{\prompts{A \textcolor{red}{baby bunny} sitting on top of a stack of pancakes.}}}
    \end{tabular}
    \end{tabular}}
    \label{subfig:mesh}
    \vspace{-0.5em}
    \caption{\textbf{Comparison with Baselines based on Stable Diffusion~\cite{rombach2022high}.} Dive3D exhibits higher quality, richer texture details, and superior alignment with human preferences, such as accurate clothing styles, and vivid fur texture.}
    \label{fig:3d_comparison}
    \vspace{-0.1em}
\end{figure*}

\begin{figure*}[!ht]
    \centering
    \setlength{\tabcolsep}{1pt}
    \setlength{\fboxrule}{1pt}
    \resizebox{0.99\textwidth}{!}{
    \begin{tabular}{c}
    \begin{tabular}{cc|cc|cccc}
        \multicolumn{2}{c}{{MVDream~\cite{shi2023mvdream}}} &
        \multicolumn{2}{c}{{DreamReward~\cite{ye2024dreamreward}}} &
        \multicolumn{4}{c}{\textbf{Dive3D (Ours)}}
        \\
        \includegraphics[width=0.15\textwidth]{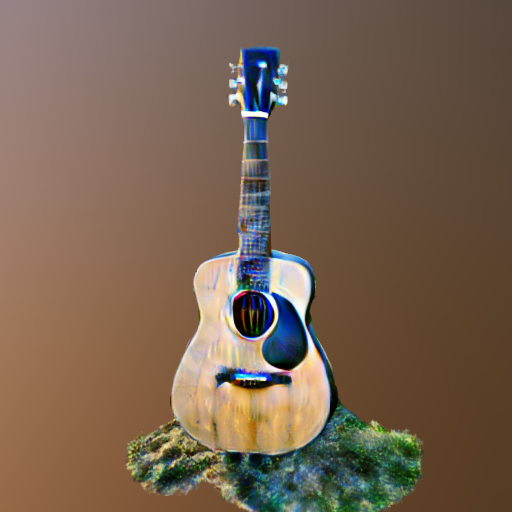} &
        \includegraphics[width=0.15\textwidth]{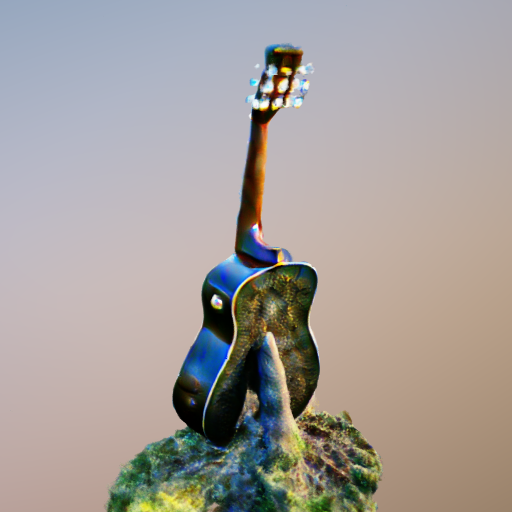} &
        \includegraphics[width=0.15\textwidth]{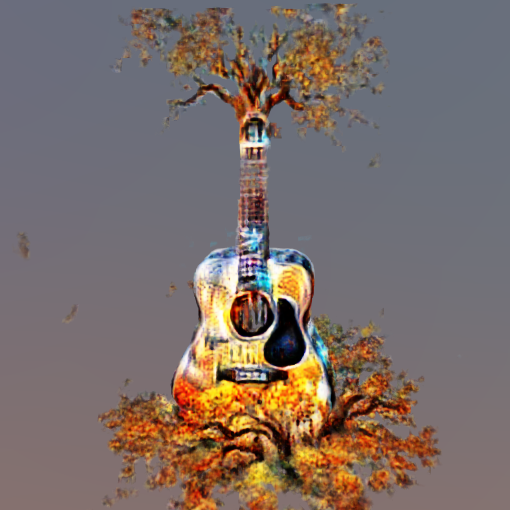} &
        \includegraphics[width=0.15\textwidth]{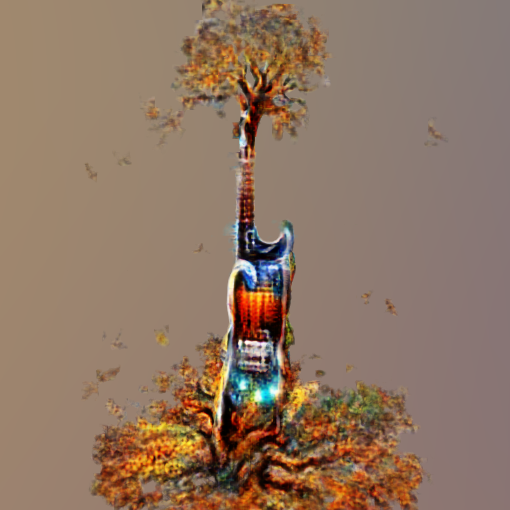} &
        \includegraphics[width=0.15\textwidth]{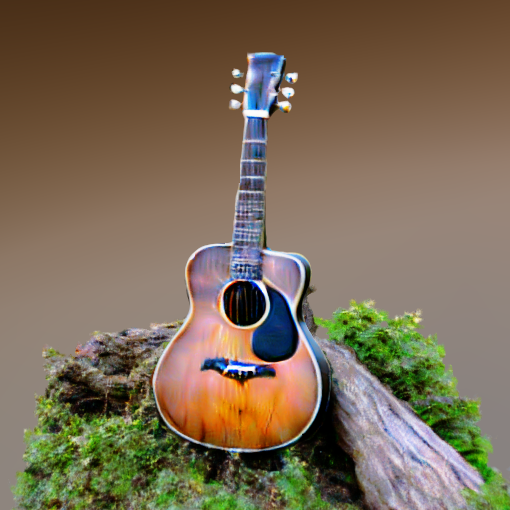} &
        \includegraphics[width=0.15\textwidth]{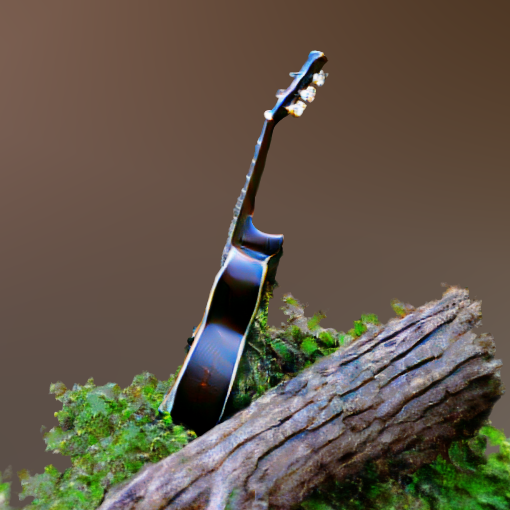} &
        \includegraphics[width=0.15\textwidth]{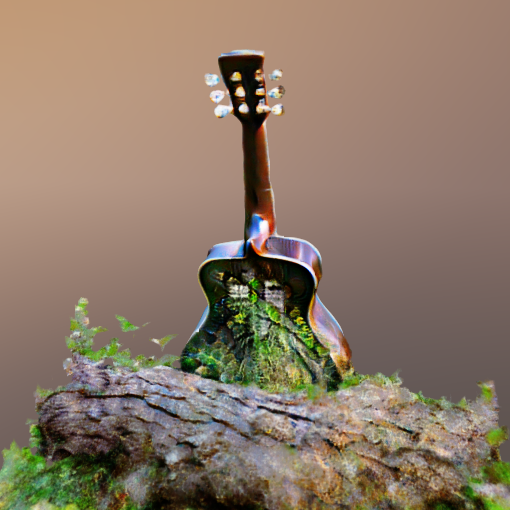} &
        \includegraphics[width=0.15\textwidth]{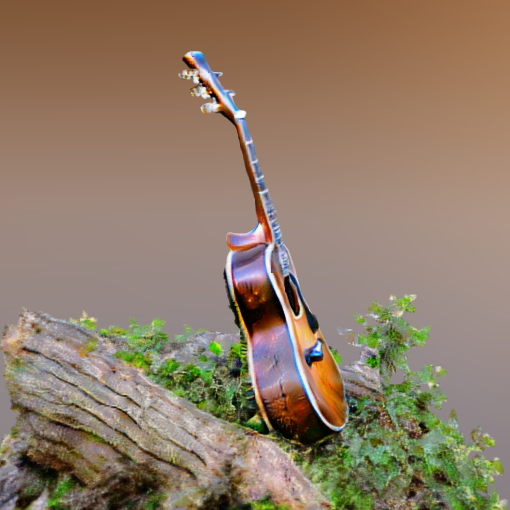} 
        \vspace{-0.5em}
        \\
        \multicolumn{8}{c}{{\prompts{A guitar resting \textcolor{red}{against an old oak tree}.}}}
        \\
        \includegraphics[width=0.15\textwidth]{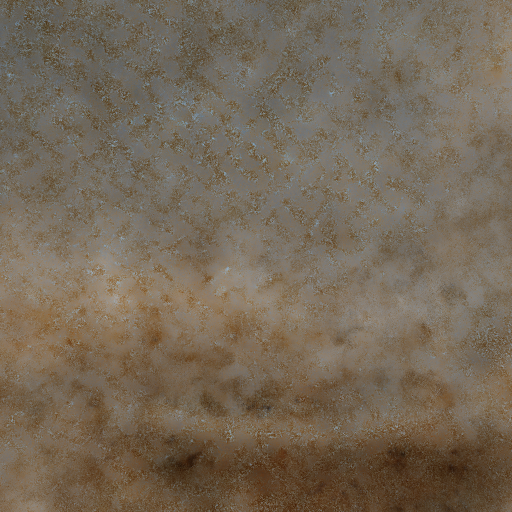} &
        \includegraphics[width=0.15\textwidth]{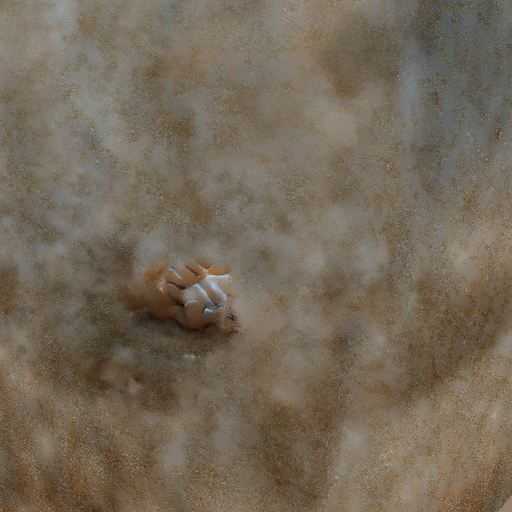} &
        \includegraphics[width=0.15\textwidth]{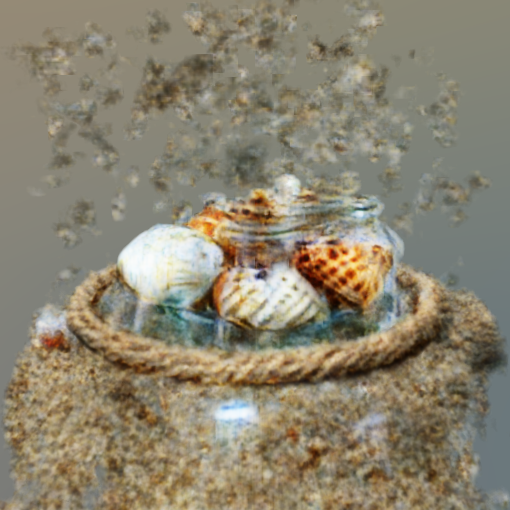} &
        \includegraphics[width=0.15\textwidth]{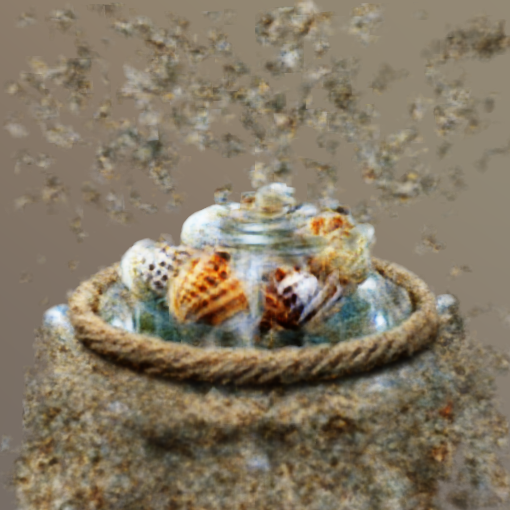} &
        \includegraphics[width=0.15\textwidth]{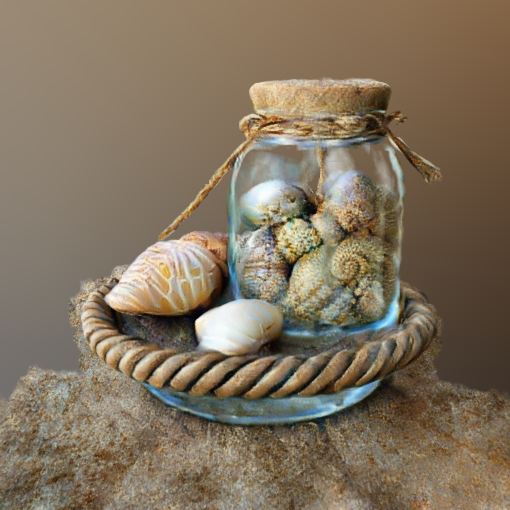} &
        \includegraphics[width=0.15\textwidth]{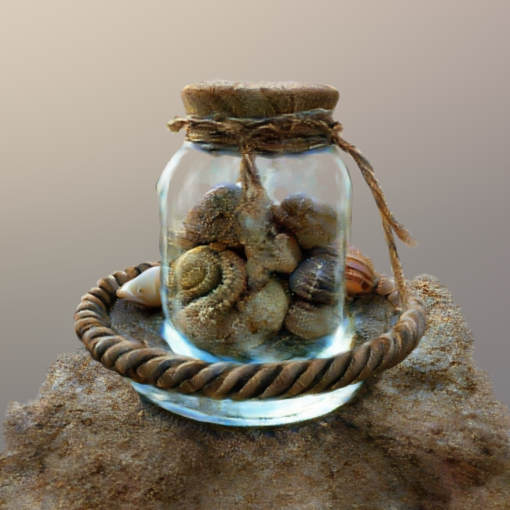} &
        \includegraphics[width=0.15\textwidth]{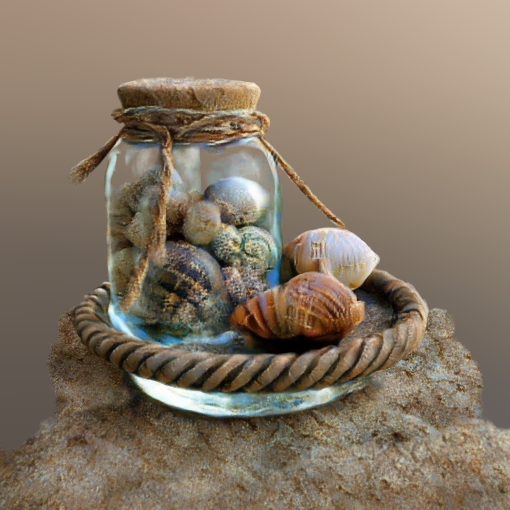} &
        \includegraphics[width=0.15\textwidth]{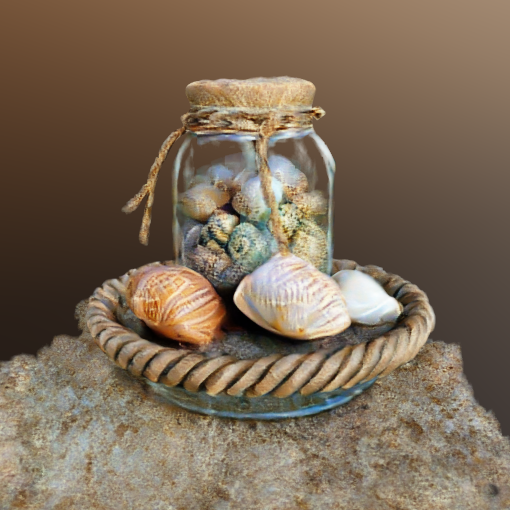} 
        \vspace{-0.5em}
        \\
        \multicolumn{8}{c}{{\prompts{Various hollow, asymmetrical, \textcolor{red}{textured seashells}, collected in a sand-filled, \textcolor{red}{clear glass jar} with a \textcolor{red}{twine-tied neck}, displayed on a windowsill.}}}
        \\
        \includegraphics[width=0.15\textwidth]{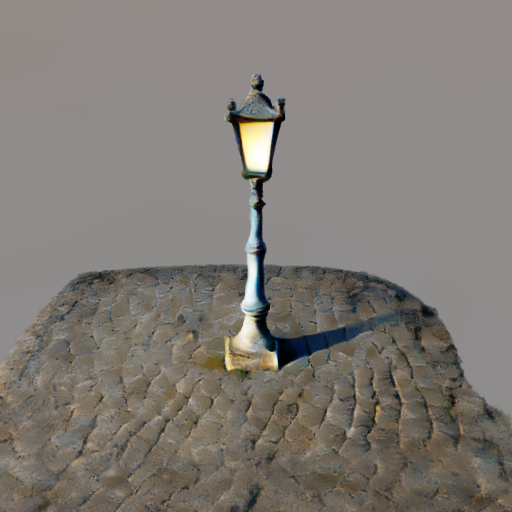} &
        \includegraphics[width=0.15\textwidth]{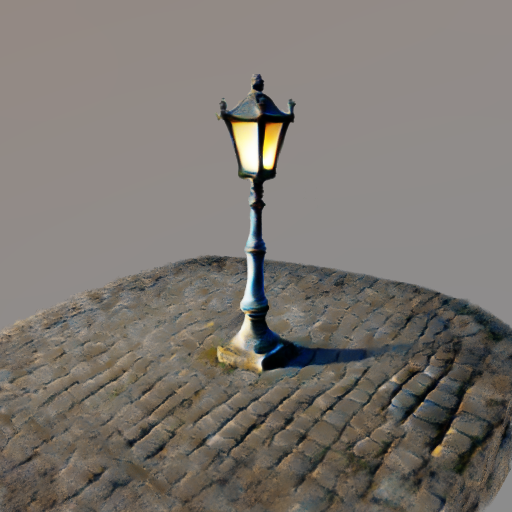} &
        \includegraphics[width=0.15\textwidth]{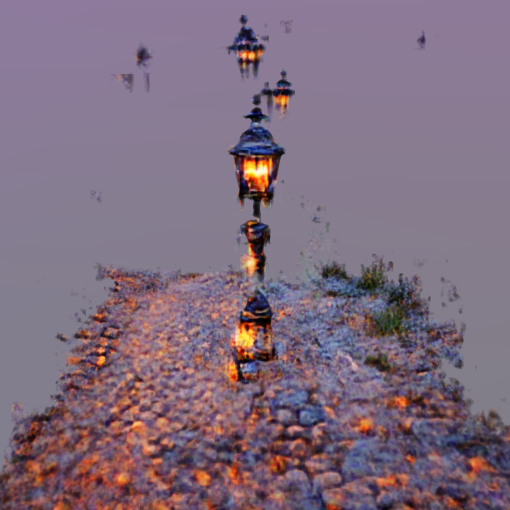} &
        \includegraphics[width=0.15\textwidth]{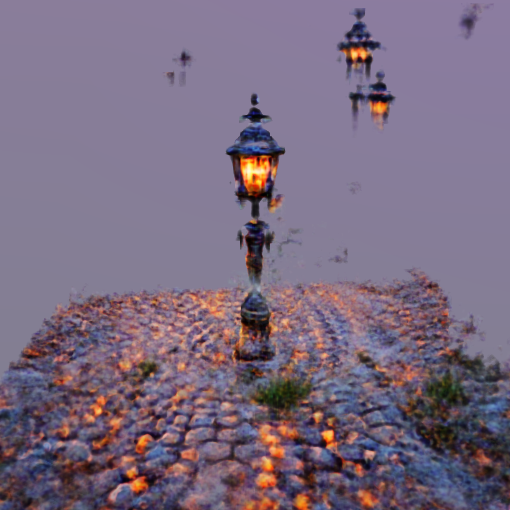} &
        \includegraphics[width=0.15\textwidth]{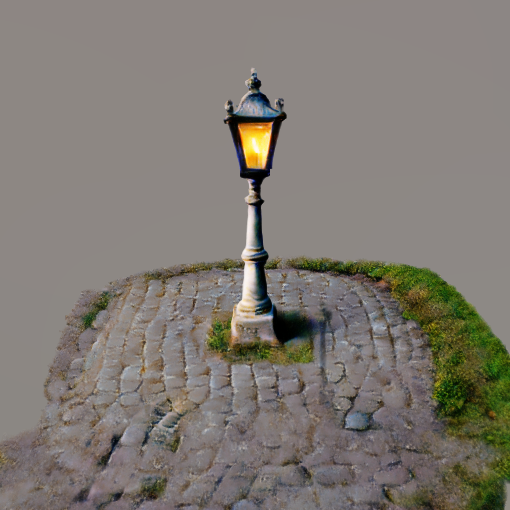} &
        \includegraphics[width=0.15\textwidth]{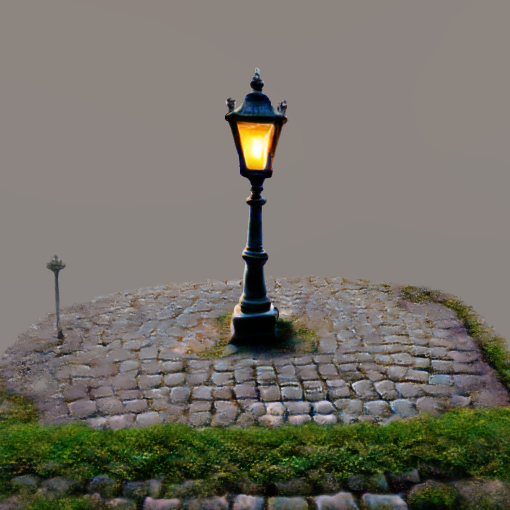} &
        \includegraphics[width=0.15\textwidth]{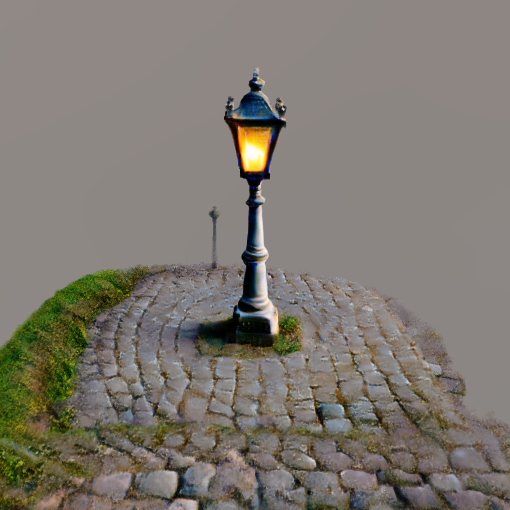} &
        \includegraphics[width=0.15\textwidth]{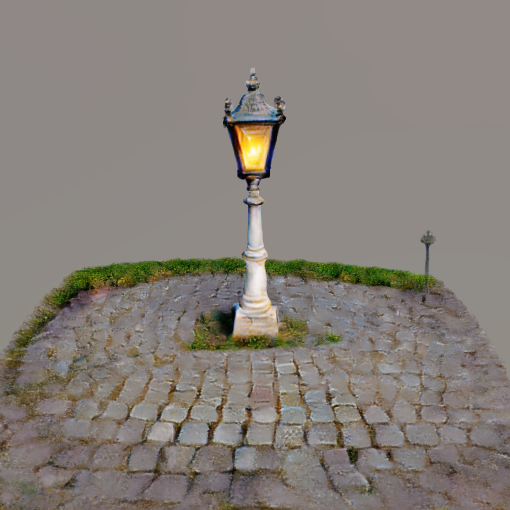} 
        \vspace{-0.5em}
        \\
        \multicolumn{8}{c}{{\prompts{A \textcolor{red}{sequence} of street lamps, casting pools of light on cobblestone paths as twilight descends.}}}
    \end{tabular}
    \end{tabular}}
    \label{subfig:mesh}
    \vspace{-0.5em}
    \caption{\textbf{Comparison with Baselines based on MVDiffusion~\cite{shi2023mvdream} and reward model~\cite{ye2024dreamreward}.} 
    Dive3D exhibits more detailed and realistic 3D generation, capturing fine-grained structures such as accurate guitar geometry and transparent glass materials.}
    \label{fig:compare-mvdream}
    \vspace{-0.1em}
\end{figure*}
\vspace{-0.1em}

\begin{figure*}[t]
    \centering
    \setlength{\tabcolsep}{1pt}
    \setlength{\fboxrule}{1pt}
    \resizebox{0.99\textwidth}{!}{
    \begin{tabular}{c}
    \begin{tabular}{ccc}
        \rotatebox{90}{\small{\,\,\,\, KL Divergence (\cite{wang2024prolificdreamer})}} &
        \includegraphics[width=0.45\textwidth]{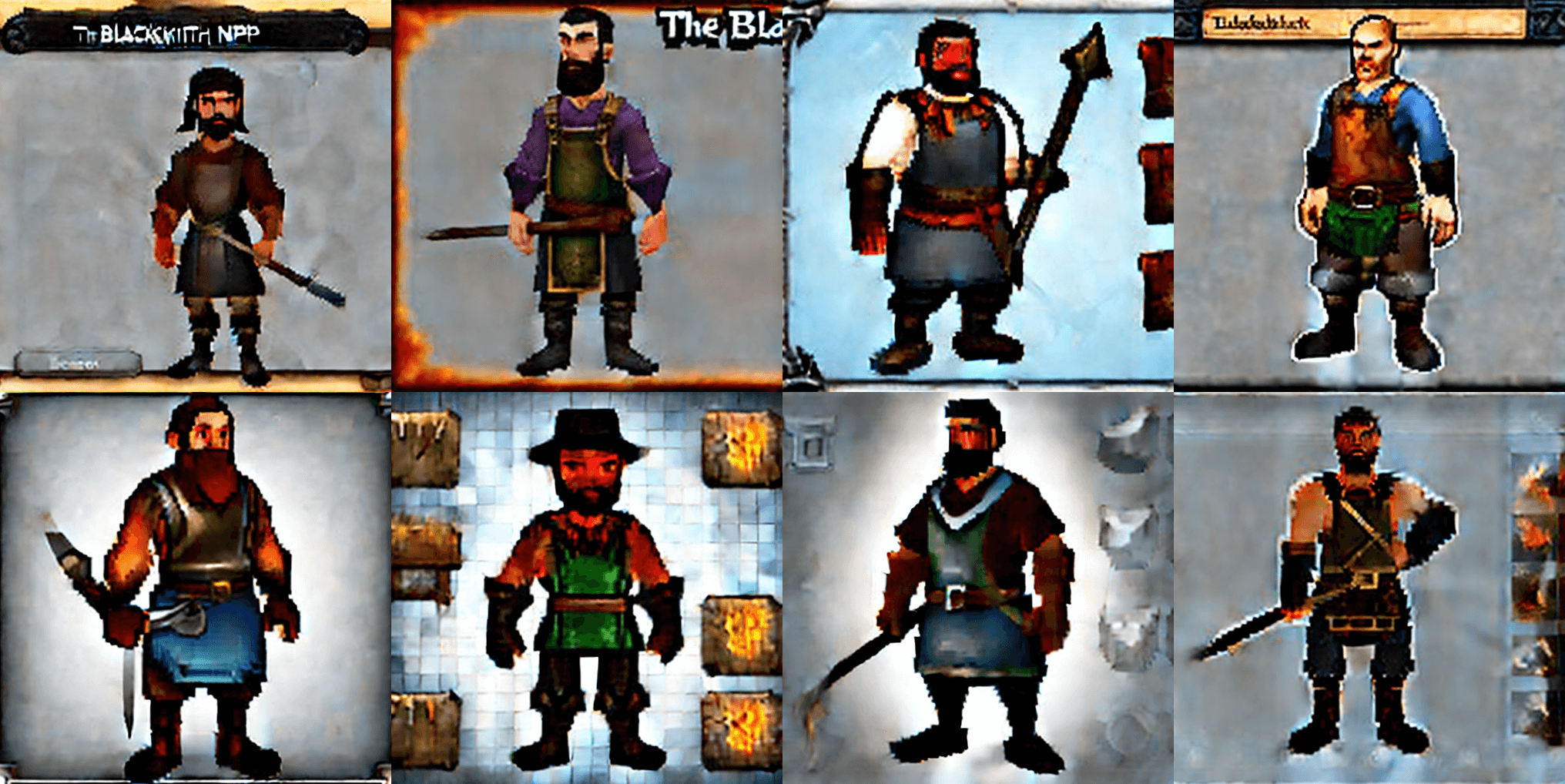} &
        \includegraphics[width=0.45\textwidth]{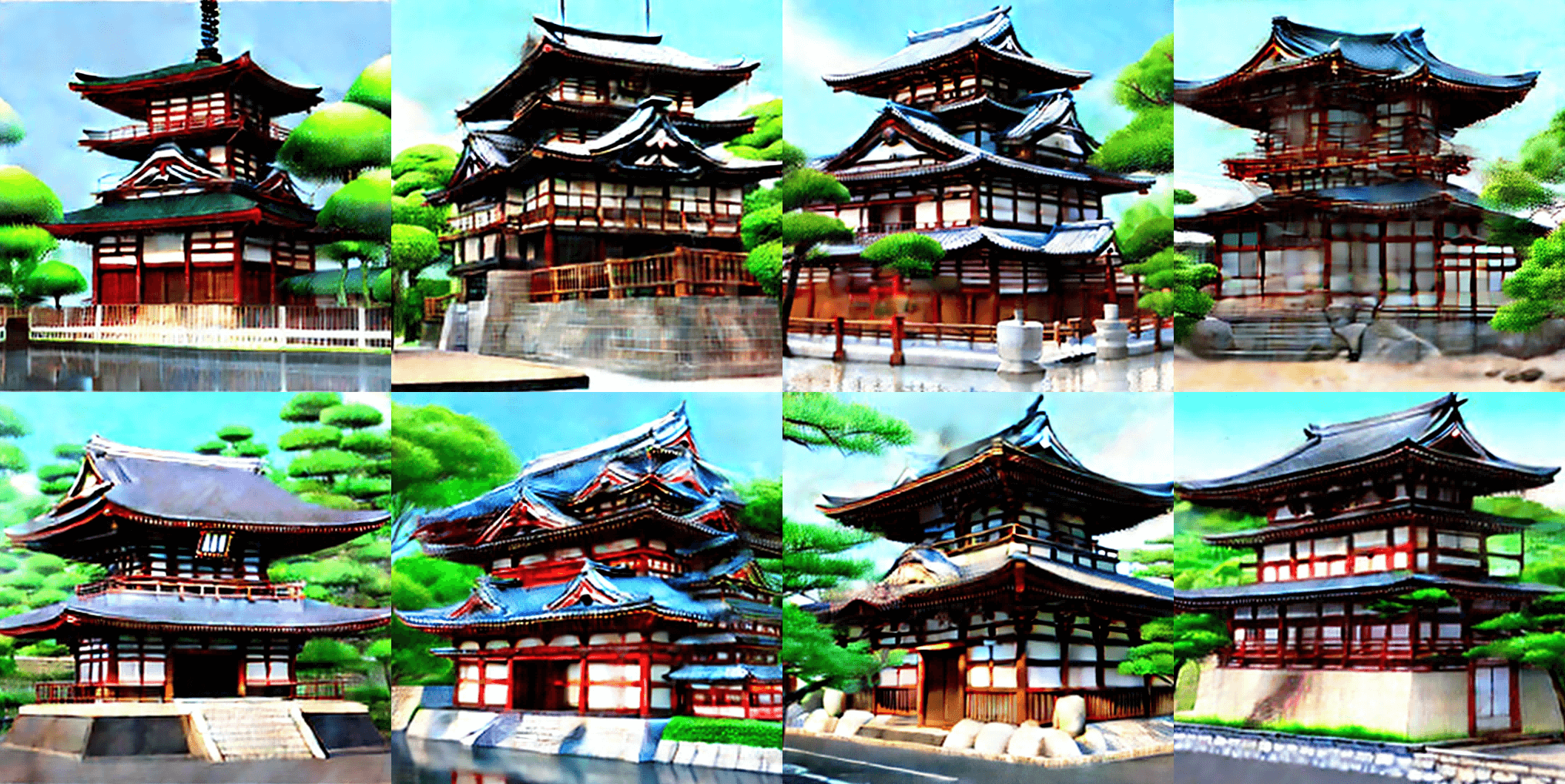} 
        \\
        \rotatebox{90}{\small{\textbf{Score Divergence (ours)}}} &
        \includegraphics[width=0.45\textwidth]{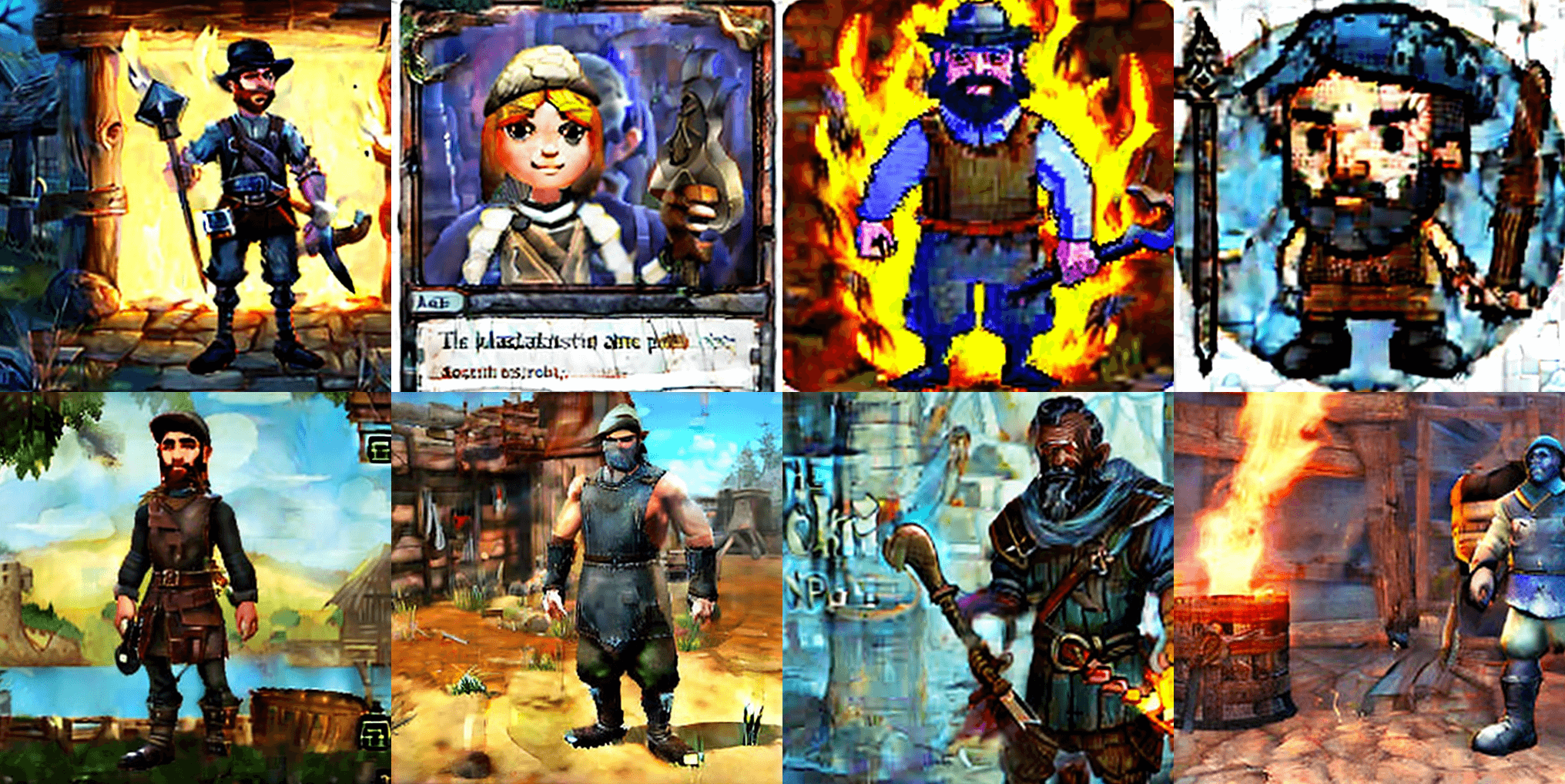}  &
        \includegraphics[width=0.45\textwidth]{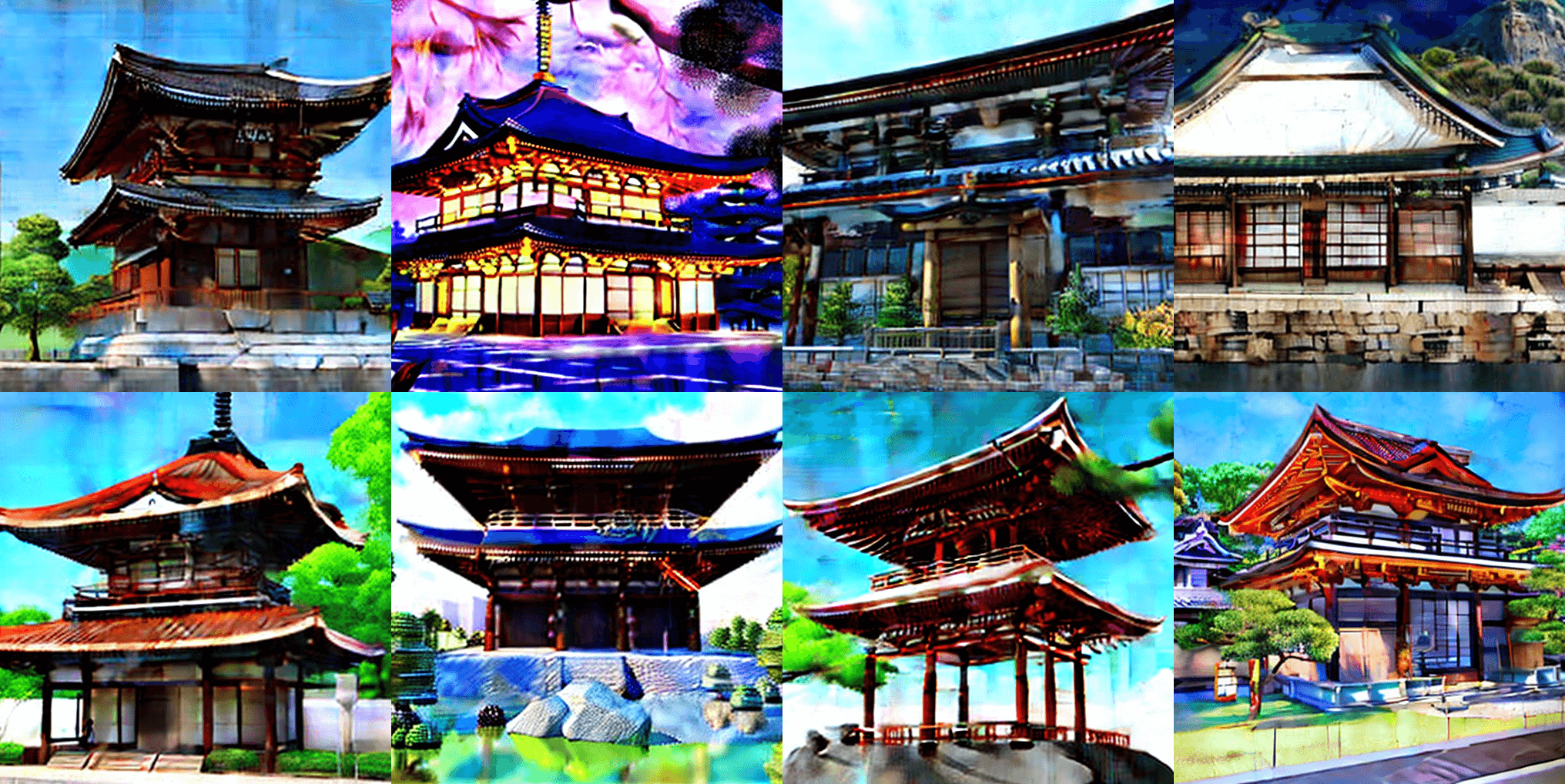}
        \vspace{-0.3em}
        \\
        &
        \multicolumn{1}{c}{{\prompts{The blacksmith NPC in the game.}}} &
        \multicolumn{1}{c}{{\prompts{A realistic Japanese building.}}}
        
    \end{tabular}
    \end{tabular}}
    \label{subfig:mesh}
    \vspace{-0.1em}
    \caption{\textbf{Score-based divergence vs. KL divergence in 2D space sampling.} The proposed score-based divergence significantly enhances the diversity of generated 2D samples, yielding more varied backgrounds and clothing in ``game character'' generation, as well as a broader range of environments, lighting conditions, and architectural features in ``Japanese building'' generation.}
    \vspace{-1em}
    \label{fig:scorevskl}
\end{figure*}

\subsection{From KL to Score-based Divergence} \label{subsec:score-divergence}
To mitigate these issues, in Dive3D we propose to replace the KL divergence with a score-based divergence, named score implict matching (SIM) loss\citep{luo2024one}, which has shown significant improvements in generation diversity in one-step diffusion and flow models\citep{luo2024one,zhou2024score,huang2024flow}. Specifically, the score-based divergence is defined between two distributions \(p\) and \(q\) as
\begin{equation}\label{eq:score_div}
D_{[0,T]}(p, q) = \int_{0}^{T} w(t) \, \mathbb{E}_{\boldsymbol{x}_t \sim \pi_t} \Big[ d\Big( s_p(\boldsymbol{x}_t) - s_q(\boldsymbol{x}_t) \Big) \Big] \, dt,
\end{equation}
where the score functions of these two distributions are given by $s_p(\boldsymbol{x}_t) = \nabla_{\boldsymbol{x}_t}\log p(\boldsymbol{x}_t) \quad \text{and} \quad s_q(\boldsymbol{x}_t) = \nabla_{\boldsymbol{x}_t}\log q(\boldsymbol{x}_t)$, $d: \mathbb{R}^d \to \mathbb{R}$ is a distance function, \(\pi_t\) is a sampling distribution whose support exceeds that of \(p_t\) and \(q_t\), and \(w(t)\) is a weighting function.
If we set $p(\cdot) = p(\boldsymbol{x}_t|y^c), p(\boldsymbol{x}_t), p_{\mathrm{ER}}(y^c, \boldsymbol{x}_t)$ and $q(\cdot) = q_\theta(\boldsymbol{x}_t|c)$, then the KL-based losses in Eqs.~\ref{eq:dive3D_loss1}-\ref{eq:dive3D_loss2} can be updated to
\begin{equation} \label{eq:score-sds-aligned}
\begin{aligned}
& \mathcal{L}_{\mathrm{Score-CDP}}(\theta) \\ &= \int_{0}^{T} w(t) \, \mathbb{E}_{\boldsymbol{x}_t \sim \pi_t} \Big[ d\Big( s_p(\boldsymbol{x}_t|y^c) - s_{q_\theta}(\boldsymbol{x}_t|c) \Big) \Big] \, dt,\\
& \mathcal{L}_{\mathrm{Score-UDP}}(\theta) \\ &= \int_{0}^{T} w(t) \, \mathbb{E}_{\boldsymbol{x}_t \sim \pi_t} \Big[ d\Big( s_p(\boldsymbol{x}_t) - s_{q_\theta}(\boldsymbol{x}_t|c) \Big) \Big] \, dt,\\
& \mathcal{L}_{\mathrm{Score-ER}}(\theta) \\ &= \int_{0}^{T} w(t) \, \mathbb{E}_{\boldsymbol{x}_t \sim \pi_t} \Big[ d\Big(\nabla_{\boldsymbol{x}_t} r\big(y^c, \hat{x}_0(\boldsymbol{x}_t)\big) - s_{q_\theta}(\boldsymbol{x}_t|c) \Big) \Big] \, dt,\\
& \mathcal{L}_{\mathrm{Dive3D}} = (1+\gamma)\mathcal{L}_{\mathrm{Score-CDP}} -\gamma \mathcal{L}_{\mathrm{Score-UDP}} + \lambda \mathcal{L}_{\mathrm{Score-ER}}
\end{aligned}
\end{equation}
This formulation offers a more effective similarity metric between the generated content and diffusion- or reward-based image distributions, yielding 3D outputs that are both more diverse and higher fidelity than those produced using traditional KL divergence.

Although this divergence may initially seem intractable, recent work \citep{luo2025one} shows that the gradient of this divergence with respect to \(\theta\) can be efficiently computed without directly differentiating the score functions by introducing a separate approximation network. For a full derivation and implementation details, please refer to Appendix.

\section{Experiment}
\label{sec:experiment}

In this section, we evaluate how our proposed score-based divergence optimization enhances both quality and diversity in text-to-3D synthesis. We perform comprehensive experiments on the GPTEval3D benchmark~\cite{wu2024gpt}, supplemented by additional 2D and 3D assessments that demonstrate the effectiveness and diversity of the method.

\subsection{Evaluation on the GPTEval3D Benchmark}

\noindent\textbf{Setup.}  
We first evaluate Dive3D on 110 creative and complex prompts from the GPTEval3D benchmark~\cite{wu2024gpt}, comparing against 9 state-of-the-art methods, including DreamFusion~\cite{poole2022dreamfusion}, DreamGaussian~\cite{tang2023dreamgaussian}, Instant3D~\cite{li2023instant3d}, Fantasia3D~\cite{chen2023fantasia3d}, Latent-NeRF~\cite{metzer2022latent}, Magic3D~\cite{lin2023magic3d}, ProlificDreamer~\cite{wang2023prolificdreamer}, MVDream~\cite{shi2023mvdream}, and DreamReward~\cite{ye2024dreamreward}. All experiments use PyTorch and the ThreeStudio framework~\cite{threestudio2023}, testing both MVDream~\cite{shi2023mvdream} and Stable Diffusion~\cite{rombach2022high} as diffusion backbones, and PickScore~\cite{kirstain2023pick} as the reward model. Optimization takes about one hour per object on a single NVIDIA A100 GPU.

\noindent\textbf{Quantitative Results.}  
Table~\ref{tab:quantitative_comparisons} reports performance of our method across six metrics, including text-asset alignment (+{53.5}), 3D plausibility (+{49}), text-geometry alignment (+{68.2}), texture details (+{67.5}), geometry details (+{35.3}), and overall performance (+{50.0}), where ``+'' indicates improvement and ``–'' indicates degradation relative to the state of the art. Dive3D achieves the top rank on every metric, demonstrating that score-based divergence guidance—especially when combined with reward models—yields substantial gains over both diffusion-only and reward-augmented baselines.

\begin{table*}[t]
  \centering
  \caption{
  \textbf{Quantitative Results on 110 Prompts from the GPTEval3D Benchmark~\cite{wu2024gpt}.} We compute all six GPTEval3D metrics—text alignment, 3D plausibility, texture–geometry coherence, geometry details,  texture details, and overall score—to comprehensively evaluate 3D generation quality. Dive3D achieves the highest score on every metric, demonstrating its superior performance.
  }
  \resizebox{\textwidth}{!}{
    \begin{tabular}{ccccccc}
    \toprule
    \multirow{2}[4]{*}{Method} & \multicolumn{6}{c}{Prompts from GPTEval3D~\cite{wu2024gpt}} \\
\cmidrule{2-7}          & Alignment & Plausibility & T-G Coherency. & Geo Details & Tex Details & Overall \\
    \midrule
    DreamFusion\cite{poole2022dreamfusion} & 1000.0  & 1000.0  & 1000.0  & 1000.0  & 1000.0  & 1000.0 \\
    DreamGaussian\cite{tang2023dreamgaussian} & 1100.6 & 953.6 & 1158.6 & 1126.2 & 1130.8 & 951.4 \\
    Fantasia3D\cite{chen2023fantasia3d} & 1067.9 & 891.9 & 1006.0 & 1109.3 & 1027.5 & 933.5 \\
    Instant3D\cite{li2023instant3d} & 1200.0  & 1087.6 & 1152.7 & 1152.0 & 1181.3 & 1097.8 \\
    Latent-NeRF\cite{metzer2022latent} & 1222.3 & 1144.8 & 1156.7 & 1180.5 & 1160.8 & 1178.7 \\
    Magic3D\cite{lin2023magic3d} & 1152.3 & 1000.8 & 1084.4 & 1178.1 & 1084.6 & 961.7 \\
    ProlificDreamer\cite{wang2023prolificdreamer} & 1261.8 & 1058.7 & 1152.0 & 1246.4 & 1180.6 & 1012.5 \\
    SyncDreamer\cite{liu2023syncdreamer} & 1041.2 & 968.8 & 1083.1 & 1064.2 & 1045.7 & 963.5 \\
    MVDream\cite{shi2023mvdream} & 1270.5 & 1147.5 & 1250.6 & 1324.9 & 1255.5 & 1097.7 \\
    DreamReward\footnotemark[1]\cite{ye2024dreamreward} & 1287.5 & 1195.0 & 1254.4 & 1295.5 & 1261.6 & 1193.3 \\
    \midrule
    DIVE3D (Ours) & \textbf{1341.0} & \textbf{1249.0} & \textbf{1322.6} & \textbf{1360.2} & \textbf{1329.1} & \textbf{1243.3} \\
    \bottomrule
    \end{tabular}%
    }
  \label{tab:quantitative_comparisons}%
  \vspace{-0.2em}
  \noindent\footnotesize\textsuperscript{1} Our metrics differ from those reported in~\cite{ye2024dreamreward} because GPT-4V has been deprecated in GPTEval3D, so we instead use GPT-4o-mini.
\end{table*}%

\noindent\textbf{Qualative Results.}  
Figure~\ref{fig:3d_comparison} compares Dive3D against methods built on Stable Diffusion (e.g., DreamFusion, Fantasia3D, ProlificDreamer), which often struggle with fine details or prompt adherence. By optimizing a score-based divergence that unifies text-conditioned diffusion priors with a differentiable reward model, Dive3D consistently produces high-fidelity, semantically precise 3D assets.

\begin{figure*}[!thb]
	\centering
	\includegraphics[width=0.85\linewidth]{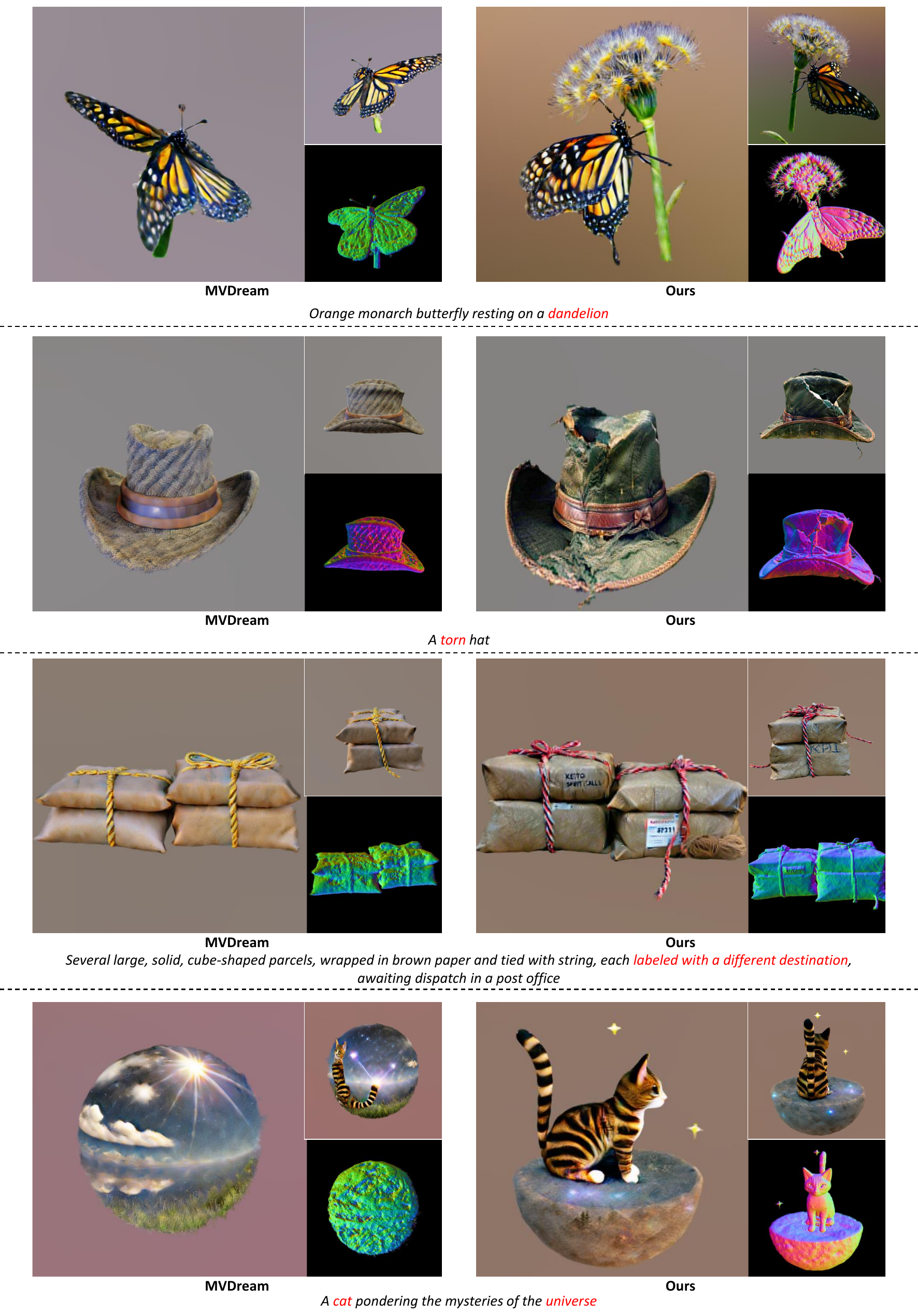}
        \vspace*{-0.2cm}
	\caption{More comparison results of Dive3D, showing improved quality over prior methods.}
	\label{fig:extended-quality}
\end{figure*}

Additional examples in Figures~\ref{fig:compare-mvdream} and~\ref{fig:extended-quality} compare Dive3D with MVDream and DreamReward. While MVDream preserves geometric consistency, it sometimes deviates from the prompt content (missing keywords highlighted in red). DreamReward improves alignment but remains constrained by its KL-based formulation and associated mode collapse. In contrast, Dive3D faithfully follows the prompt, delivers rich detail and appealing aesthetics, and maintains strong visual coherence.

\subsection{Analysis on Generation Diversity}
\noindent\textbf{Setup.}
We then show that score-based divergences produce more diverse, information-rich outputs than traditional KL-based losses. To evaluate this, we test our method in both 2D and 3D settings—using Stable Diffusion~\cite{rombach2022high} as the backbone. In 2D, we represent scenes with 2D Neural Radiance Fields; in 3D, we use full 3D NeRFs. We primarily compare against ProlificDreamer~\cite{wang2023prolificdreamer}, the leading KL-divergence–based method that leverages variational score distillation (VSD) to maximize diversity in text-to-3D generation. On a single NVIDIA A100 GPU, our 2D experiments complete in roughly 30 minutes, while the 3D evaluations take about 9 hours.

\noindent\textbf{2D Results.}
We begin by evaluating 2D generation, where we distill a 2D neural field from a text-to-image diffusion model. This task shares the same mathematical formulation as our text-to-3D problem but is computationally less demanding because it does not involve camera poses.
As shown in Fig.~\ref{fig:scorevskl}, for both game character and realistic architecture generation tasks, the score-based divergence consistently produces more diverse samples than KL divergence. For instance, when generating ``a realistic Japanese building,'' the KL-based method consistently generates towers with standard color schemes (predominantly red and blue), uniform backgrounds (lush green trees), and similar weather and time conditions (sunny daytime). In contrast, the score-based approach generates outputs with varied lighting (e.g., night scenes, snowy settings) and diverse architectural features (e.g., towers, pavilions, and residential houses). A similar trend is observed in the game character generation task: while the KL-based SDS loss tends to produce similar archetypes, the score-based loss reveals a wider range of characters, clothing styles, and backgrounds.

\begin{figure*}[!thb]
	\centering
	\includegraphics[width=0.91\linewidth]{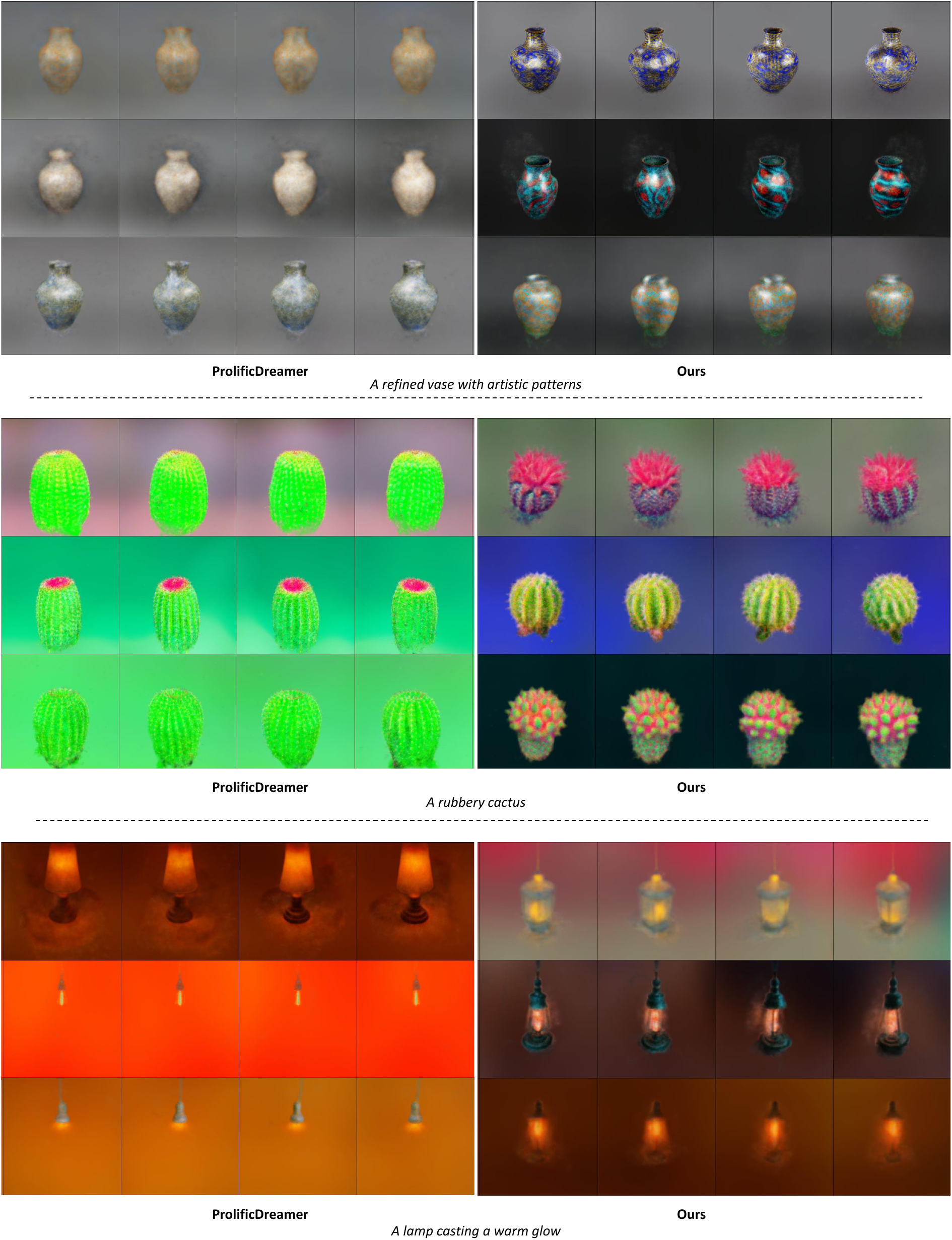}
        \vspace*{-0.2cm}
	\caption{More comparison results of Dive3D, showing improved diversity over prior methods.}
	\label{fig:extended-diversity}
\end{figure*}

\noindent\textbf{3D Results.}
These diversity gains naturally and effectively generalize to 3D synthesis. Figure~\ref{fig:teaser}(a) compares the output for “a pirate ship in the sky” under the KL-based VSD loss versus our score-based divergence. As expected, our approach produces a far wider range of geometric shapes, surface textures, and background scenes—from bright sunny skies to dark thunderous clouds. Figure \ref{fig:extended-diversity} offers additional examples across diverse prompts to reinforce this finding, illustrating how score-based divergence yields richer variation in colors, object styles, material properties, and environmental details.

\section{Conclusion}
\label{sec:conclusion}
In this paper, we present Dive3D, a novel framework that boosts both diffusion-based distillation and reward-guided optimization by substituting their asymmetric KL-divergence objectives with a powerful score-based divergence. Evaluated on the GPTEval3D and other benchmarks, Dive3D effectively alleviates mode collapse, yielding substantially greater diversity while simultaneously improving text alignment, geometric plausibility, and visual fidelity.

\noindent\textbf{Limitations and Future Work.}
Although Dive3D delivers compelling results, its runtime remains slower than recent LRM-based methods. In future work, we plan to integrate our score-based divergence with latent reconstruction models by first distilling diverse, text-driven multi-view generators and then combining them with LRM techniques to achieve rapid, high-fidelity, and diverse 3D synthesis. 

\newpage
{
    \small
    \bibliographystyle{ieeenat_fullname}
    \bibliography{main}
}

\newpage
\appendix
\clearpage

\onecolumn

\begin{center}
  {\Large \bfseries Supplementary Material \par} 
  \vspace{1.5em} 
\end{center}

\setcounter{page}{1}
\renewcommand{\thesection}{\Alph{section}}
\setcounter{section}{0}

\section{Derivation of the Score-based Loss Function}
\label{sec: appendix-theorem}
The derivation is based on the so-called score-projection identity which was first found
in~\cite{vincent2011connection} to bridge denoising score matching and denoising auto-encoders. Later the identity is applied by~\cite{zhou2024score, luo2025one} for deriving distillation methods for image generation. We appreciate the efforts of~\cite{zhou2024score} and introduce the conditional score-projection identity for inverse problems here:

\begin{theorem}\label{thm:example}
(Conditional score-projection identity).
Let $\boldsymbol{u}(\cdot, \theta)$ be a vector-valued function, under mild conditions, the identity holds:
\end{theorem}

\begin{equation}
    \mathbb{E}_{\substack{\boldsymbol{x}_0\sim p_{\theta}(\boldsymbol{x}_0|\boldsymbol{y}) \\ \boldsymbol{x}_t|\boldsymbol{x}_0\sim q_t(\boldsymbol{x}_t|\boldsymbol{x}_0)}} \boldsymbol{u}(\boldsymbol{x}_t,\boldsymbol{y}, \theta)^T\left\{\boldsymbol{s}_{p_{\theta,t}}(\boldsymbol{x}_t|\boldsymbol{y})-\nabla_{\boldsymbol{x}_t} \log q_t(\boldsymbol{x}_t|\boldsymbol{x}_0)\right\}=0, \quad \forall \theta.
\end{equation}

Next, we turn to derive the loss function.

\textit{Proof.} We prove a more general result. Let $\boldsymbol{u}(\cdot, \theta)$ be a vector-valued function, the so-called score-projection identity~\cite{vincent2011connection} holds,

\begin{equation}
    \mathbb{E}_{\substack{\boldsymbol{x}_0\sim p_{\theta}(\boldsymbol{x}_0|\boldsymbol{y}) \\ \boldsymbol{x}_t|\boldsymbol{x}_0\sim q_t(\boldsymbol{x}_t|\boldsymbol{x}_0)}} \boldsymbol{u}(\boldsymbol{x}_t,\boldsymbol{y}, \theta)^T\left\{\boldsymbol{s}_{p_{\theta,t}}(\boldsymbol{x}_t|\boldsymbol{y})-\nabla_{\boldsymbol{x}_t} \log q_t(\boldsymbol{x}_t|\boldsymbol{x}_0)\right\}=0, \quad \forall \theta.
\label{equ:a1}
\end{equation}

Taking $\theta$ gradient on both sides of identity~\ref{equ:a1}, we have:

\begin{equation}
\begin{array}{l}
    0=\mathbb{E}_{\substack{\boldsymbol{x}_{0} \sim p_{\theta}(\boldsymbol{x}_0|\boldsymbol{y}) \\
    \boldsymbol{x}_{t} \mid \boldsymbol{x}_{0} \sim q_{t}\left(\boldsymbol{x}_{t} \mid \boldsymbol{x}_{0}\right)}} \frac{\partial}{\partial \boldsymbol{x}_{t}}\left\{\boldsymbol{u}\left(\boldsymbol{x}_{t}, \boldsymbol{y}, \theta\right)^{T}\left\{\boldsymbol{s}_{p_{\theta, t}}\left(\boldsymbol{x}_{t}|\boldsymbol{y}\right)-\nabla_{\boldsymbol{x}_{t}} \log q_{t}\left(\boldsymbol{x}_{t} \mid \boldsymbol{x}_{0}\right)\right\}\right\} \frac{\partial \boldsymbol{x}_{t}}{\partial \theta} \\
    +\mathbb{E}_{\substack{\boldsymbol{x}_{0} \sim p_{\theta}(\boldsymbol{x}_0|\boldsymbol{y}) \\
    \boldsymbol{x}_{t} \mid \boldsymbol{x}_{0} \sim q_{t}\left(\boldsymbol{x}_{t} \mid \boldsymbol{x}_{0}\right)}} \frac{\partial}{\partial \boldsymbol{x}_{0}}\left\{\boldsymbol{u}\left(\boldsymbol{x}_{t}, \boldsymbol{y},\theta\right)^{T}\left\{-\nabla_{\boldsymbol{x}_{t}} \log q_{t}\left(\boldsymbol{x}_{t} \mid \boldsymbol{x}_{0}\right)\right\}\right\} \frac{\partial \boldsymbol{x}_{0}}{\partial \theta} \\
    +\mathbb{E}_{\substack{\boldsymbol{x}_{0} \sim p_{\theta}(\boldsymbol{x}_0|\boldsymbol{y}) \\
    \boldsymbol{x}_{t} \mid \boldsymbol{x}_{0} \sim q_{t}\left(\boldsymbol{x}_{t} \mid \boldsymbol{x}_{0}\right)}}\boldsymbol{u}\left(\boldsymbol{x}_{t}, \boldsymbol{y},\theta\right)^{T} \frac{\partial}{\partial \theta}\left\{\boldsymbol{s}_{p_{\theta, t}}\left(\boldsymbol{x}_{t}|\boldsymbol{y}\right)\right\}+\frac{\partial}{\partial \theta} \boldsymbol{u}\left(\boldsymbol{x}_{t}, \boldsymbol{y},\theta\right)^{T} \boldsymbol{s}_{p_{\theta,t}}\left(\boldsymbol{x}_{t}|\boldsymbol{y}\right) \\
    =\mathbb{E}_{\substack{\boldsymbol{x}_{0} \sim p_{\theta}(\boldsymbol{x}_0|\boldsymbol{y}) \\
    \boldsymbol{x}_{t} \mid \boldsymbol{x}_{0} \sim q_{t}\left(\boldsymbol{x}_{t} \mid \boldsymbol{x}_{0}\right)}} \boldsymbol{u}\left(\boldsymbol{x}_{t}, \boldsymbol{y},\theta\right)^{T} \frac{\partial}{\partial \theta}\left\{\boldsymbol{s}_{p_{\theta, t}}\left(\boldsymbol{x}_{t}|\boldsymbol{y}\right)\right\} \\
    +\mathbb{E}_{\substack{\boldsymbol{x}_{0} \sim p_{\theta}(\boldsymbol{x}_0|\boldsymbol{y}) \\
    \boldsymbol{x}_{t} \mid \boldsymbol{x}_{0} \sim q_{t}\left(\boldsymbol{x}_{t} \mid \boldsymbol{x}_{0}\right)}}\left\{\frac{\partial}{\partial \boldsymbol{x}_{t}}\left\{\boldsymbol{u}\left(\boldsymbol{x}_{t}, \boldsymbol{y}, \theta\right)^{T}\left\{\boldsymbol{s}_{p_{\theta, t}}\left(\boldsymbol{x}_{t}|\boldsymbol{y}\right)-\nabla_{\boldsymbol{x}_{t}} \log q_{t}\left(\boldsymbol{x}_{t}| \boldsymbol{x}_{0}\right)\right\}\right\} \frac{\partial \boldsymbol{x}_{t}}{\partial \theta}\right. \\
    +\frac{\partial}{\partial \boldsymbol{x}_{0}}\left\{\boldsymbol{u}\left(\boldsymbol{x}_{t}, \boldsymbol{y}, \theta\right)^{T}\left\{-\nabla_{\boldsymbol{x}_{t}} \log q_{t}\left(\boldsymbol{x}_{t}|\boldsymbol{x}_{0}\right)\right\}\right\} \frac{\partial \boldsymbol{x}_{0}}{\partial \theta}
    \left.+\frac{\partial}{\partial \theta} \boldsymbol{u}\left(\boldsymbol{x}_{t}, \boldsymbol{y}, \theta\right)^{T} \boldsymbol{s}_{p_{\theta,t}}\left(\boldsymbol{x}_{t}|\boldsymbol{y}\right)\right\} \\
    =\mathbb{E}_{\boldsymbol{x}_{t} \sim p_{\theta, t}} \boldsymbol{u}\left(\boldsymbol{x}_{t},\boldsymbol{y}, \theta\right)^{T} \frac{\partial}{\partial \theta}\left\{\boldsymbol{s}_{p_{\theta, t}}\left(\boldsymbol{x}_{t}|\boldsymbol{y}\right)\right\} 
    +\frac{\partial}{\partial \theta} \mathbb{E}_{\substack{\boldsymbol{x}_{0} \sim p_{\theta, t}}} \boldsymbol{u}\left(\boldsymbol{x}_{t}, \boldsymbol{y},\theta\right)^{T}\left\{\boldsymbol{s}_{p_{\mathrm{sg}[\theta], t}}\left(\boldsymbol{x}_{t}|\boldsymbol{y}\right)-\nabla_{\boldsymbol{x}_{t}} \log q_{t}\left(\boldsymbol{x}_{t}|\boldsymbol{x}_{0}\right)\right\}
\end{array}
\end{equation}

Therefore we have the following identity:
\begin{equation}
    \mathbb{E}_{\boldsymbol{x}_{t} \sim p_{\theta, t}} \boldsymbol{u}\left(\boldsymbol{x}_{t},\boldsymbol{y}, \theta\right)^{T} \frac{\partial}{\partial \theta}\left\{\boldsymbol{s}_{p_{\theta, t}}\left(\boldsymbol{x}_{t}|\boldsymbol{y}\right)\right\} 
    =-\frac{\partial}{\partial \theta} \mathbb{E}_{\substack{\boldsymbol{x}_{0} \sim p_{\theta, t}}} \boldsymbol{u}\left(\boldsymbol{x}_{t}, \boldsymbol{y},\theta\right)^{T}\left\{\boldsymbol{s}_{p_{\mathrm{sg}[\theta], t}}\left(\boldsymbol{x}_{t}|\boldsymbol{y}\right)-\nabla_{\boldsymbol{x}_{t}} \log q_{t}\left(\boldsymbol{x}_{t}|\boldsymbol{x}_{0}\right)\right\}
\end{equation}

which holds for arbitrary function $\boldsymbol{u}(\cdot, \theta)$ and parameter $\theta$. If we set
\begin{equation}
\begin{array}{l}
    \boldsymbol{u}(\boldsymbol{x}_t, \boldsymbol{y}, \theta)= \mathbf{d}'(\boldsymbol{k}_t) \\
    \boldsymbol{k}_t= \boldsymbol{s}_{p_{\mathrm{sg}[\theta], t}}\left(\boldsymbol{x}_{t}|\boldsymbol{y}\right)-s_{q_{t}}\left(\boldsymbol{x}_{t}\right)-s_{q_{t}}\left(\boldsymbol{y}|\boldsymbol{x}_{t}\right)
\end{array}
\end{equation}

Then we formally have the loss function as:
\begin{equation}
    \begin{array}{l}
loss
=\mathbb{E}_{\boldsymbol{x}_{t} \sim p_{\theta, t}}\left\{-\mathbf{d}^{\prime}\left(\boldsymbol{k}_{t}\right)\right\}^{T}\left\{\boldsymbol{s}_{p_{\mathrm{sg}[\theta], t}}\left(\boldsymbol{x}_{t}|\boldsymbol{y}\right)-\nabla_{\boldsymbol{x}_{t}} \log q_{t}\left(\boldsymbol{x}_{t}|\boldsymbol{x}_{0}\right)\right\}
\end{array}
\end{equation}

\begin{figure*}[!ht]
    \centering
    \setlength{\tabcolsep}{1pt}
    \setlength{\fboxrule}{1pt}
    \resizebox{0.99\textwidth}{!}{
    \begin{tabular}{c}
    \begin{tabular}{cccccccc}
        \multicolumn{2}{c}{{DreamGaussian~\cite{tang2023dreamgaussian}}} &
        \multicolumn{2}{c}{{GaussianDreamer~\cite{yi2024gaussiandreamer}}} &
        \multicolumn{2}{c}{{Dive3D (Ours), Particle 1}} &
        \multicolumn{2}{c}{{Dive3D (Ours), Particle 2}}
        \\
        \includegraphics[width=0.15\textwidth]{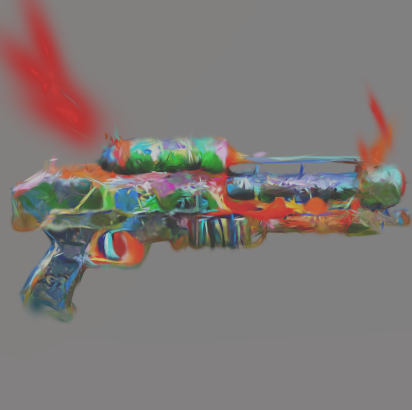} &
        \includegraphics[width=0.15\textwidth]{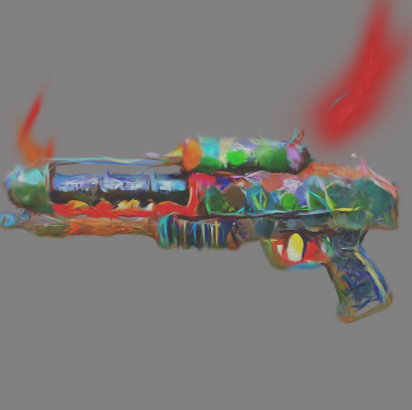} &
        \includegraphics[width=0.15\textwidth]{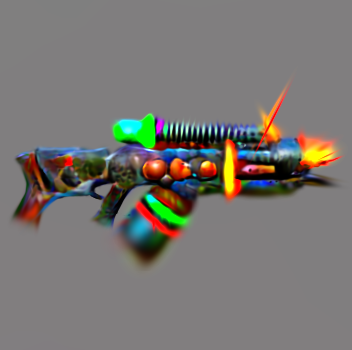} &
        \includegraphics[width=0.15\textwidth]{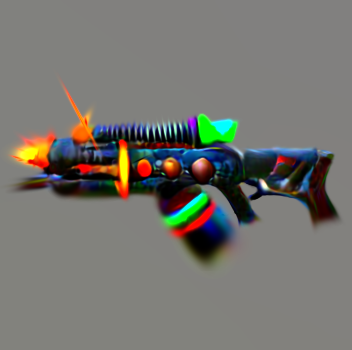} &
        \includegraphics[width=0.15\textwidth]{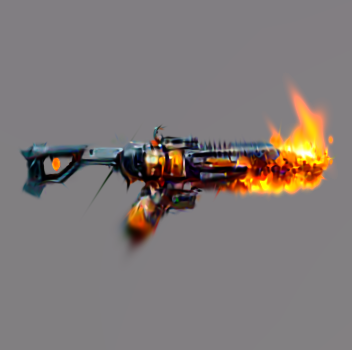} &
        \includegraphics[width=0.15\textwidth]{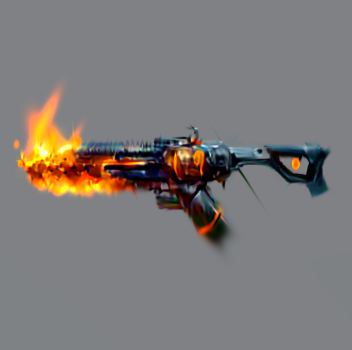} &
        \includegraphics[width=0.15\textwidth]{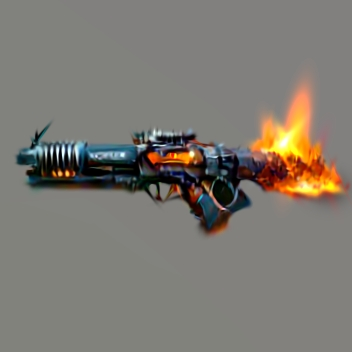} &
        \includegraphics[width=0.15\textwidth]{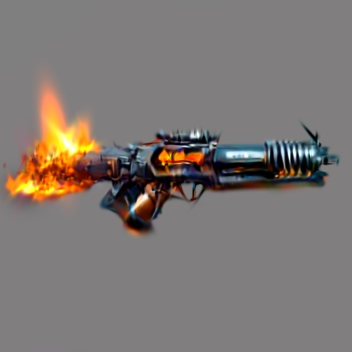}
        \vspace{-0.5em}
        \\
        \multicolumn{8}{c}{{\prompts{Flamethrower, with fire, scifi, cyberpunk...}}}
        \\
        \includegraphics[width=0.15\textwidth]{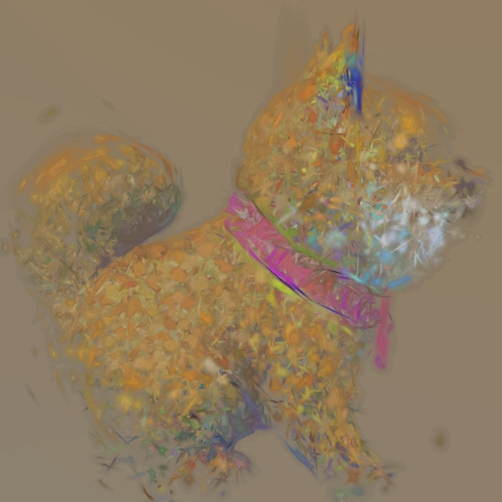} &
        \includegraphics[width=0.15\textwidth]{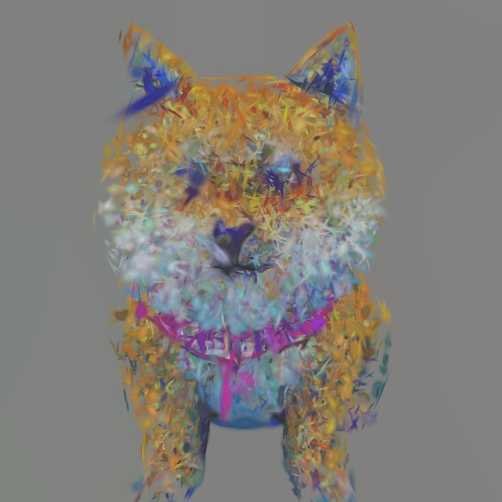} &
        \includegraphics[width=0.15\textwidth]{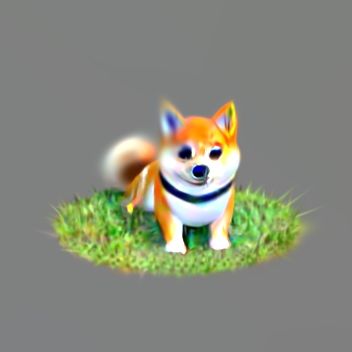} &
        \includegraphics[width=0.15\textwidth]{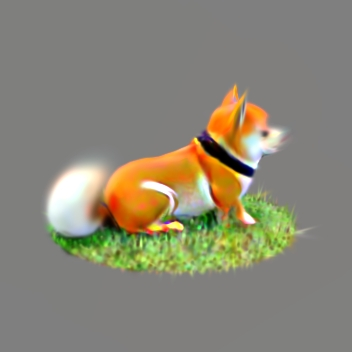} &
        \includegraphics[width=0.15\textwidth]{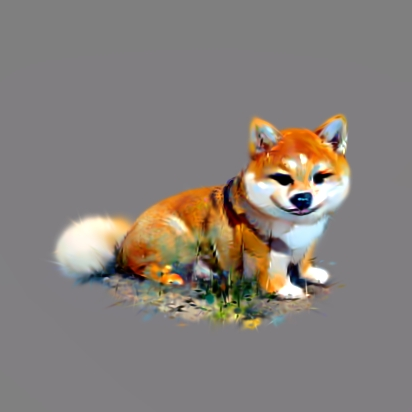} &
        \includegraphics[width=0.15\textwidth]{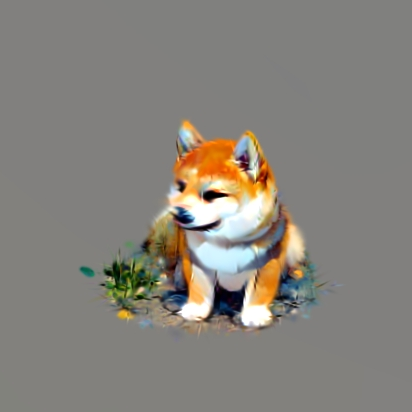} &
        \includegraphics[width=0.15\textwidth]{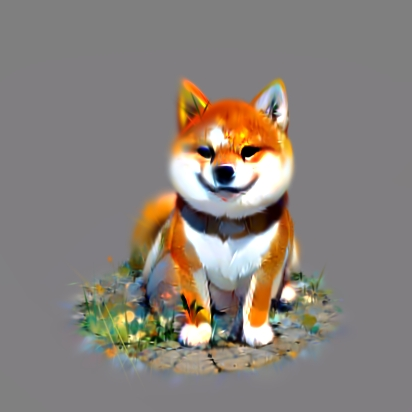} &
        \includegraphics[width=0.15\textwidth]{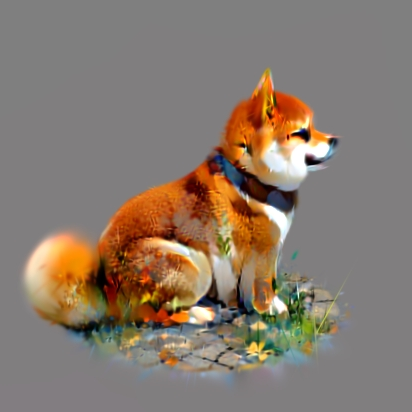} 
        \vspace{-0.5em}
        \\
        \multicolumn{8}{c}{{\prompts{A cute Shiba Inu is sitting on the ground.}}} \\
    \end{tabular}
    \end{tabular}}
    \label{subfig:mesh}
    \vspace{-1em}
    \caption{\textbf{Comparison with baselines on Gaussian Splatting.} Dive3D can generate diverse and human-preferred 3D Gaussian splattings.}
    \label{fig:comparisiongsalone}
\end{figure*}

\begin{figure*}[!ht]
    \centering
    \setlength{\tabcolsep}{1pt}
    \setlength{\fboxrule}{1pt}
    \begin{tabular}{@{}ll@{}}
        \rotatebox{90}{\small{Gaussian Splatting}} &
        \includegraphics[width=0.98\textwidth]{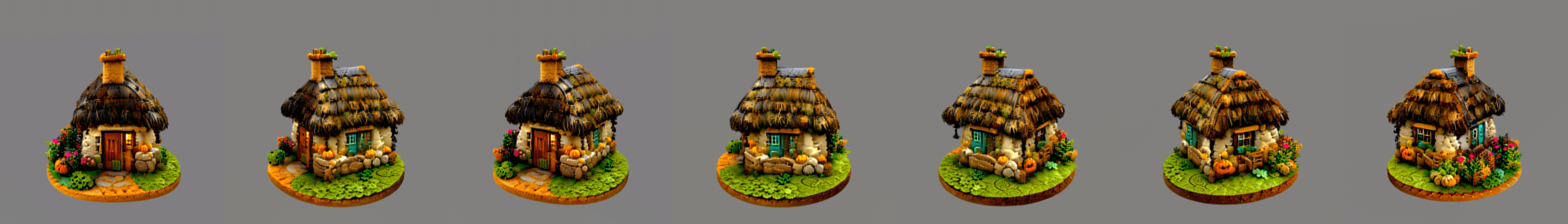}
        \vspace{-0.1em}
        \\
        \rotatebox{90}{\small{\,\,\,\,\,\,\,\,\,\,\,\,\,\,\,\,\,\,NeRF}} &
        \includegraphics[width=0.98\textwidth]{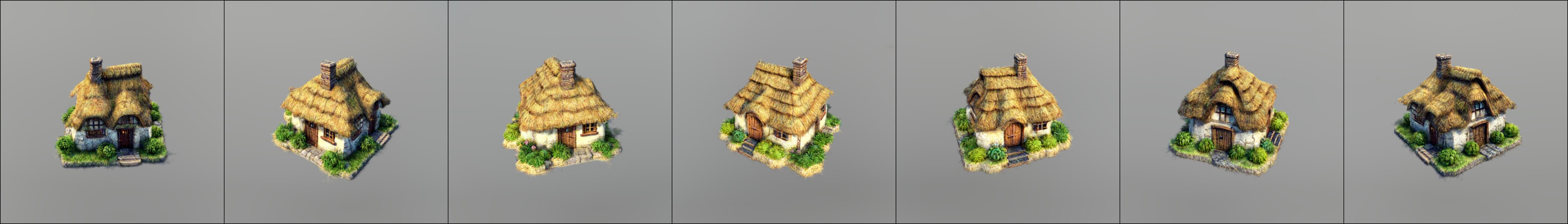} 
        \vspace{-0.1em}
        \\
        \rotatebox{90}{\small{\,\,\,\,\,\,\,\,\,\,\,\,\,\,\,\,\,\,Mesh}} &
        \includegraphics[width=0.98\textwidth]{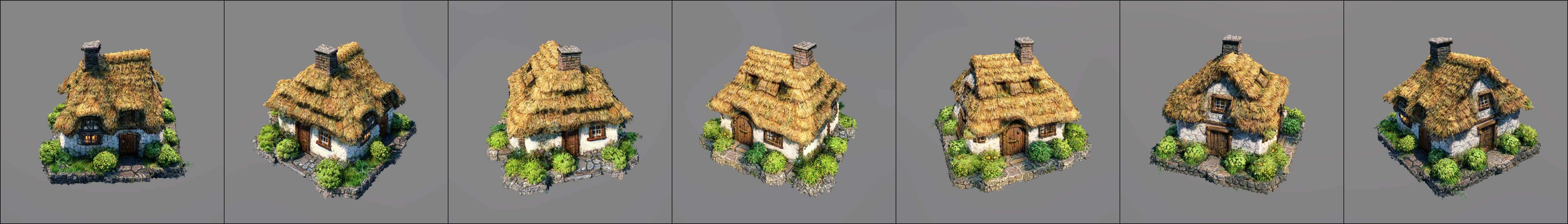} 
        \\
        \rotatebox{90}{\small{Gaussian Splatting}} &
        \includegraphics[width=0.98\textwidth]{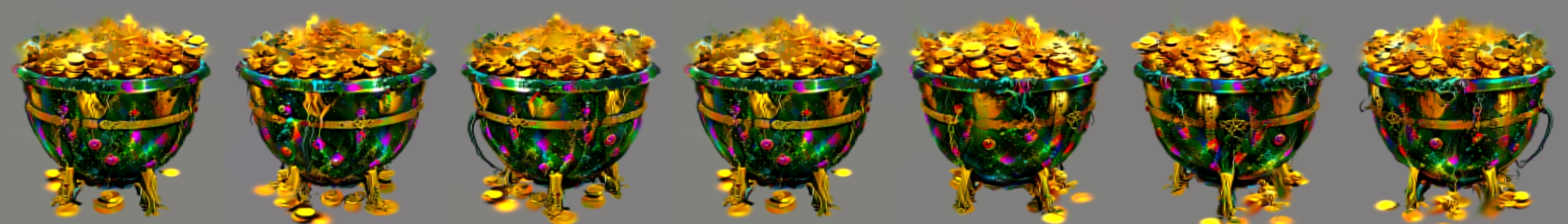}
        \vspace{-0.1em}
        \\
        \rotatebox{90}{\small{\,\,\,\,\,\,\,\,\,\,\,\,\,\,\,\,\,\,NeRF}} &
        \includegraphics[width=0.98\textwidth]{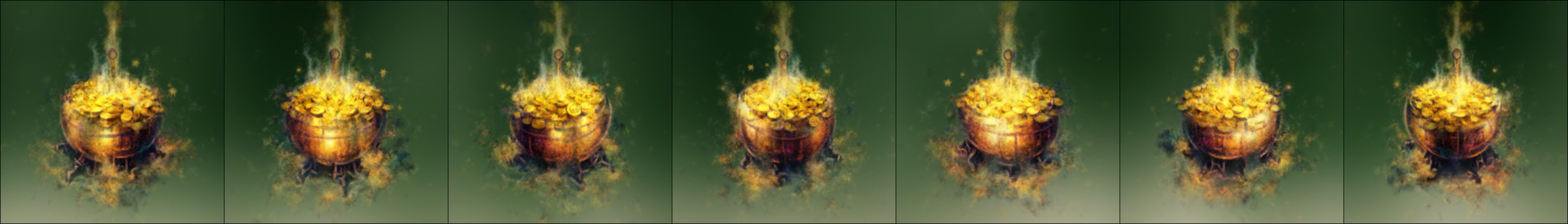} 
        \vspace{-0.1em}
        \\
        \rotatebox{90}{\small{\,\,\,\,\,\,\,\,\,\,\,\,\,\,\,\,\,\,Mesh}} &
        \includegraphics[width=0.98\textwidth]{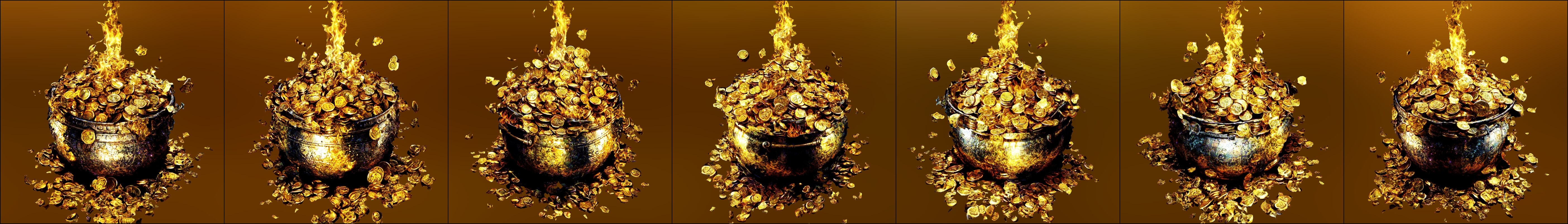} \\
    \end{tabular}
    \label{subfig:mesh}
    \vspace{-0.3em}
    \caption{\textbf{Dive3D can generate 3D objects across different 3D representations.} We show the results on Gaussian splitting, NeRF and Mesh. The prompts are \prompts{A 3D model of an adorable cottage with a thatched roof} and \prompts{A cauldron full of gold coins.}}
    \label{fig:comparisiongs}
\end{figure*}

\section{Implementation Details}

\paragraph{Pseudo-code for Dive3D}
A detailed pseudo-code for Dive3D is presented in algorithm~\ref{alg:dive3d}.

\begin{algorithm}[t]
\caption{Pseudo-code for Dive3D}
\label{alg:dive3d}
\begin{tabular}{@{}r@{\hspace{1em}}p{0.85\linewidth}@{}} 

1 & \textbf{Input:} Text-to-image diffusion score network $s_{\phi}(\cdot, \cdot, \cdot)$ and its LoRA parameters $\varphi$, Reward function $R(\cdot, \cdot)$ \\
2 & \textbf{Input:} Text prompt $y$, Differentiable 3D renderer $g(\theta, v)$ \\
3 & \textbf{Input:} Learning rate $\eta$ for 3D parameters $\theta$, Timestep weighting function $w(t)$ \\
4 & \textbf{Input:} Hyperparameters: $\gamma$ (guidance scale coefficient), $\lambda$ (reward weight) \\
5 & \textbf{Input:} Noise schedule parameters $\alpha_t, \sigma_t$ \\
6 & \\

7 & \textbf{Initialization:} Randomly initialize 3D representation parameters $\theta_0$ \\
8 & \\

9 & \textbf{for} each training step $i=1, \dots, N$ \textbf{do} \\
10 & \hspace{1em} Sample camera viewpoint $v$ \\
11 & \hspace{1em} Render 2D image from current 3D model: $x_0 \leftarrow g(\theta, v)$ \\
12 & \hspace{1em} Sample timestep $t \sim \text{Uniform}(0, T)$ and noise $\epsilon \sim \mathcal{N}(0, I)$ \\
13 & \hspace{1em} Compute noisy image: $x_t \leftarrow \alpha_t x_0 + \sigma_t \epsilon$ \\
14 & \hspace{1em} Compute score of the current rendered data distribution: $s_{\text{rendered}} \leftarrow -\epsilon/\sigma_t$ \\
15 & \\

\multicolumn{2}{@{}l}{\hspace{1em}\textbf{--- Score-based Conditional Diffusion Prior (S-CDP) Gradient ---}} \\
16 & \hspace{1em} Estimate conditional target score: $s_{\text{cond\_target}} \leftarrow s_{\phi}(x_t, y, t)$ \\
17 & \hspace{1em} Compute S-CDP gradient: $\nabla_{\theta}\mathcal{L}_{\text{S-CDP}} \approx w(t) d_{\text{score}}(s_{\text{cond\_target}}, s_{\varphi}(x_t, y, t))$ \\
18 & \\

\multicolumn{2}{@{}l}{\hspace{1em}\textbf{--- Score-based Unconditional Diffusion Prior (S-UDP) Gradient ---}} \\
19 & \hspace{1em} Estimate unconditional target score: $s_{\text{uncond\_target}} \leftarrow s_{\phi}(x_t, \emptyset, t)$ \\
20 & \hspace{1em} Compute S-UDP gradient: $\nabla_{\theta}\mathcal{L}_{\text{S-UDP}} \approx w(t) d_{\text{score}}(s_{\text{uncond\_target}}, s_{\varphi}(x_t, y, t))$ \\
21 & \\

\multicolumn{2}{@{}l}{\hspace{1em}\textbf{--- Score-based Explicit Reward (S-ER) Gradient ---}} \\
22 & \hspace{1em} Estimate denoised image $\hat{x}_0(x_t)$ \\
23 & \hspace{1em} Estimate target reward score: $s_{\text{reward\_target}} \leftarrow \nabla_{x_t} R(y, \hat{x}_0(x_t))$ \\
24 & \hspace{1em} Compute S-ER gradient: $\nabla_{\theta}\mathcal{L}_{\text{S-ER}} \approx w(t) d_{\text{score}}(s_{\text{reward\_target}}, s_{\varphi}(x_t, y, t))$ \\
25 & \\

\multicolumn{2}{@{}l}{\hspace{1em}\textbf{--- Total Loss Gradient and Parameter Update ---}} \\
26 & \hspace{1em} Compute total Dive3D loss gradient (based on Eq. 16): \\
27 & \hspace{1em} $\nabla_{\theta}\mathcal{L}_{\text{Dive3D}} \leftarrow (1+\gamma)\nabla_{\theta}\mathcal{L}_{\text{S-CDP}} - \gamma\nabla_{\theta}\mathcal{L}_{\text{S-UDP}} + \lambda\nabla_{\theta}\mathcal{L}_{\text{S-ER}}$ \\
28 & \hspace{1em} Update parameters: $\theta \leftarrow \theta - \eta \nabla_{\theta}\mathcal{L}_{\text{Dive3D}}$ \\
29 & \hspace{1em} Update parameters: $\varphi \leftarrow \text{Denoising Score Matching}(x_t, t)$ \\
30 & \textbf{end for} \\

\end{tabular}
\end{algorithm}

\textbf{Backbones.}
We utilize Stable Diffusion~\cite{rombach2022high} v2.1 and the MVDream model~\cite{shi2023mvdream} sd-v2.1-base-4view. 
To produce diverse 3D representations, we adapt our algorithm to generate NeRF outputs and follow the geometry and mesh refinement stages as outlined in ProlificDreamer\cite{wang2024prolificdreamer}. 
Additionally, we employ the Gaussian Splatting generation pipeline from GaussianDreamer\cite{yi2024gaussiandreamer} for 3D Gaussian Splatting outputs. 
To align the generated 3D content with human preferences, we utilize the PickScore model~\cite{kirstain2023pick} and also compare the results with those from the established ImageReward model~\cite{xu2023imagereward}.

\textbf{Hyperparameters.} 
we use a CFG scale of 7.5 for mesh generation and 20 for Gaussian Splatting generation. We observed that a high CFG scale tends to oversaturate colors, while a low CFG scale hinders object convergence. With algorithmic improvements, we found that a smaller CFG scale can produce high-quality 3D objects with smooth colors.
For PickScore, we use a scale of 100 for mesh generation and 10,000 for Gaussian Splatting generation. 
Empirically, we found that a high PickScore scale enhances visual richness but may lead to color oversaturation. 
The optimal PickScore scale ranges between 50–200 for mesh generation and 5,000–10,000 for Gaussian Splatting generation.

For our quantitative evaluations, we maintain consistent experimental setups across all methods, employing similar pipelines that include 3D representation selection, training procedures, shape initialization, and the use of a teacher diffusion model. 
We uniformly apply identical negative prompts during the assessments. 
The primary distinction between the methods lies in their respective loss functions.

The total training time is approximately 20 minutes on an RTX 3090 GPU for Gaussian Splatting generation, 1.5 hours on an A100 GPU for NeRF generation and 12 hours for mesh generation.

\section{Impact of Divergence Loss Combinations}
In this section, we investigate how different combinations of text-to-3D divergence losses affect generation performance. Our framework involves three fundamental divergence losses: the CDP loss, the UDP loss, and the ER loss. 
\begin{equation}
\mathcal{L}_{\mathrm{Dive3D}} = \alpha_1 \mathcal{L}_{\mathrm{CDP}}- \alpha_2 \mathcal{L}_{\mathrm{UDP}} + \alpha_3 \mathcal{L}_{\mathrm{ER}}
\end{equation}
We systematically explore the effects of different loss weights by analyzing all possible paired combinations.
\vspace{-0.5em}

\begin{figure*}[!ht]
    \centering
    \setlength{\tabcolsep}{1pt}
    \setlength{\fboxrule}{1pt} 
    \begin{subfigure}[t]{\textwidth}
        \centering
    \setlength{\tabcolsep}{1pt}
    \setlength{\fboxrule}{1pt}
    \resizebox{0.97\textwidth}{!}{
    \begin{tabular}{c}
    \begin{tabular}{cccccccc}
        $\alpha_{2}/\alpha_1=0$ & \multicolumn{6}{c}{\begin{tikzpicture}[baseline]  
                    \draw[->, >=latex, line width=0.35mm] (0,0.1) -- (14cm,0.1);  
                    \end{tikzpicture}  } & $\alpha_{2}/\alpha_1=2$
        \\
        \begin{overpic}[width=0.13\linewidth]{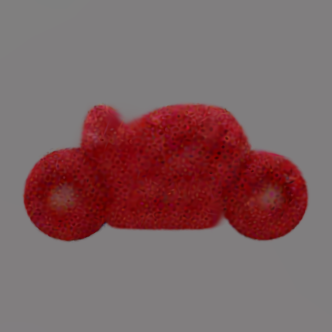}
					\end{overpic} &
        \begin{overpic}[width=0.13\linewidth]{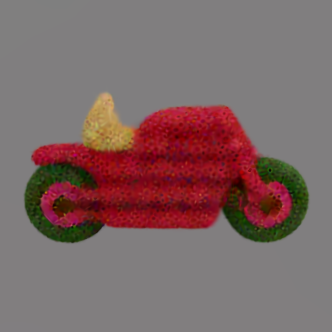}
					\end{overpic} &
        \begin{overpic}[width=0.13\linewidth]{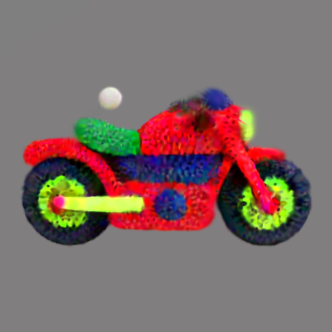}
					\end{overpic} &
        \begin{overpic}[width=0.13\linewidth]{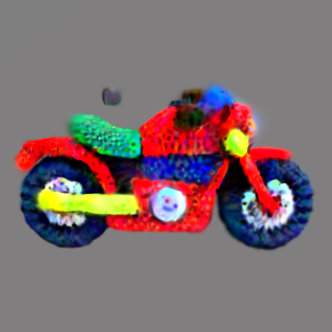}
					\end{overpic} &
        \begin{overpic}[width=0.13\linewidth]{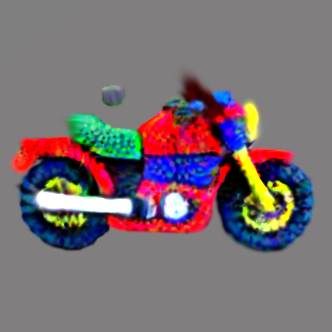}
					\end{overpic} &
        \begin{overpic}[width=0.13\linewidth]{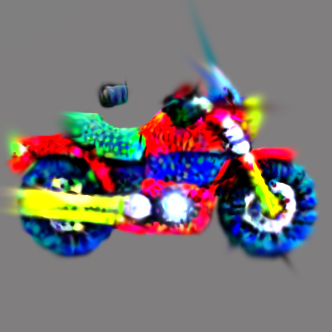}
					\end{overpic} &
        \begin{overpic}[width=0.13\linewidth]{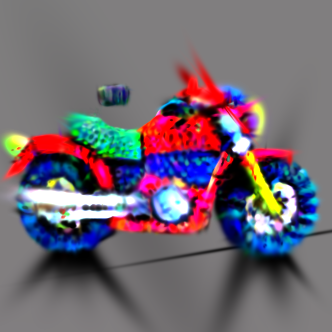}
					\end{overpic} &
        \begin{overpic}[width=0.13\linewidth]{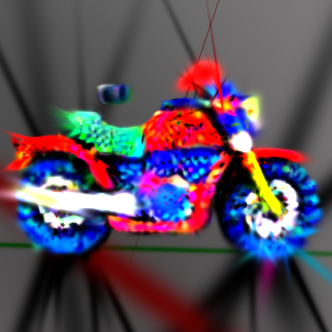}
					\end{overpic}
        \vspace{-0.2em}
        \\
        \multicolumn{8}{c}{{(a) $\mathcal{L}=\alpha_1 \mathcal{L}_{\mathrm{CDP}} - \alpha_2 \mathcal{L}_{\mathrm{UDP}}$. The prompt is: \prompts{An amigurumi motorcycle.}}}
        \end{tabular}
        \vspace{-0.2em}
    \end{tabular}}
        \label{subfig:nerf}
    \label{subfig:sds-udp}
    \end{subfigure}
    \vspace{0.5em}
    
    \begin{subfigure}[t]{\textwidth}
        \centering
    \setlength{\tabcolsep}{1pt}
    \setlength{\fboxrule}{1pt}
    \resizebox{1\textwidth}{!}{
    \begin{tabular}{c}
    \begin{tabular}{ccccccccc}
        & $\alpha_3/\alpha_1=0$ & \multicolumn{6}{c}{\begin{tikzpicture}[baseline]  
                    \draw[->, >=latex, line width=0.35mm] (0,0.1) -- (14cm,0.1);  
                    \end{tikzpicture}  } & $\alpha_3/\alpha_1=1e5$
        \\
        \rotatebox{90}{\small{ImageReward}} &
        \begin{overpic}[width=0.13\linewidth]{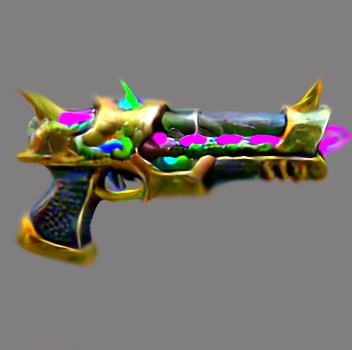}
						\put(65,67){{\includegraphics[width=0.04\linewidth]{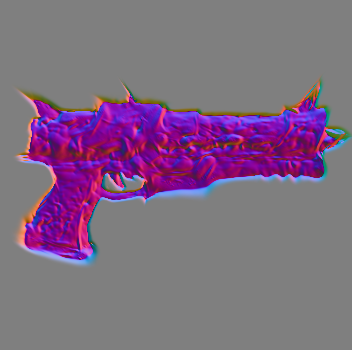}}}
					\end{overpic} &
        \begin{overpic}[width=0.13\linewidth]{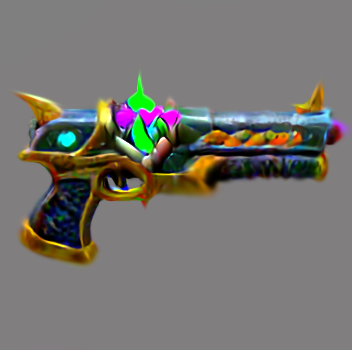}
						\put(65,67){{\includegraphics[width=0.04\linewidth]{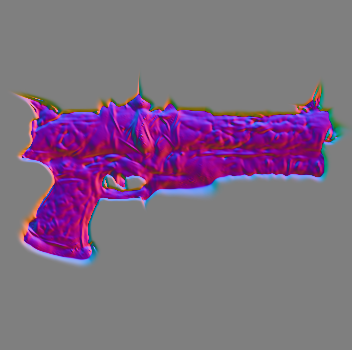}}}
					\end{overpic} &
        \begin{overpic}[width=0.13\linewidth]{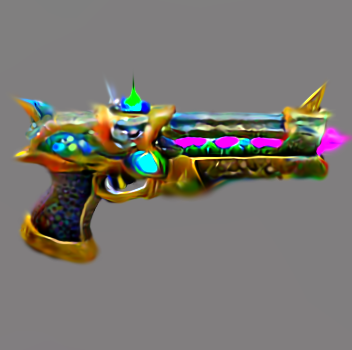}
						\put(65,67){{\includegraphics[width=0.04\linewidth]{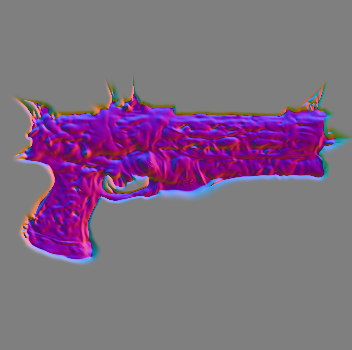}}}
					\end{overpic} &
        \begin{overpic}[width=0.13\linewidth]{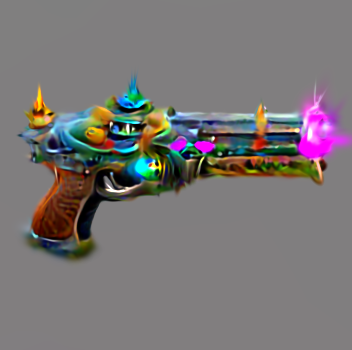}
						\put(65,67){{\includegraphics[width=0.04\linewidth]{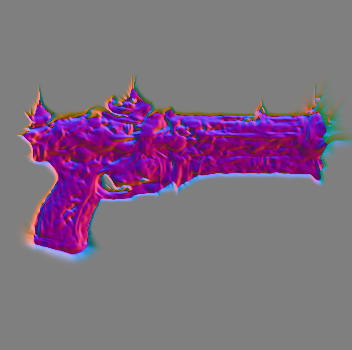}}}
					\end{overpic} &
        \begin{overpic}[width=0.13\linewidth]{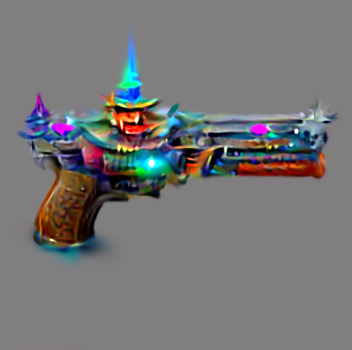}
						\put(65,67){{\includegraphics[width=0.04\linewidth]{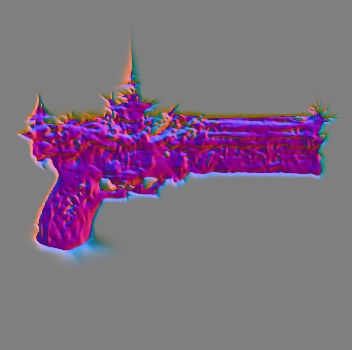}}}
					\end{overpic} &
        \begin{overpic}[width=0.13\linewidth]{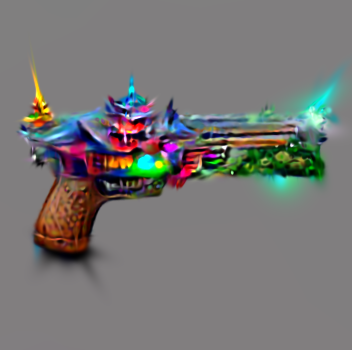}
						\put(65,67){{\includegraphics[width=0.04\linewidth]{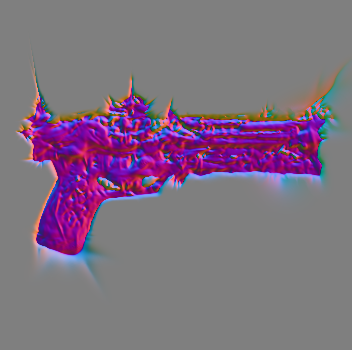}}}
					\end{overpic} &
        \begin{overpic}[width=0.13\linewidth]{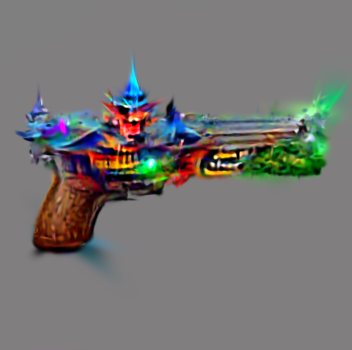}
						\put(65,67){{\includegraphics[width=0.04\linewidth]{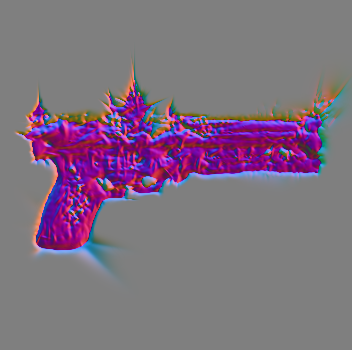}}}
					\end{overpic} &
        \begin{overpic}[width=0.13\linewidth]{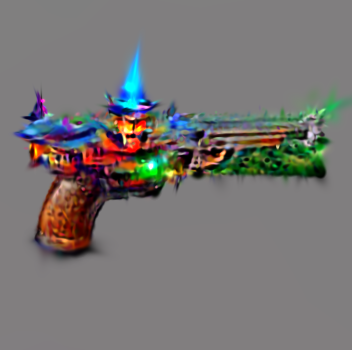}
						\put(65,67){{\includegraphics[width=0.04\linewidth]{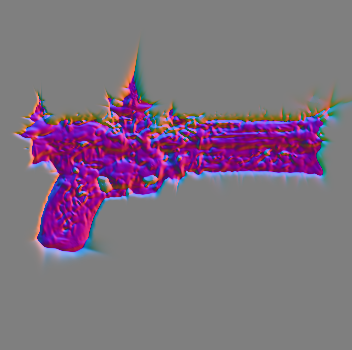}}}
					\end{overpic}
        \vspace{-0.2em}
        \\
        \rotatebox{90}{\,\,\,\small{PickScore}} &
        \begin{overpic}[width=0.13\linewidth]{./figure/figure4/ir1_row1_col1.png}
						\put(65,67){{\includegraphics[width=0.04\linewidth]{./figure/figure4/ir1_row1_col2.png}}}
					\end{overpic} &
        \begin{overpic}[width=0.13\linewidth]{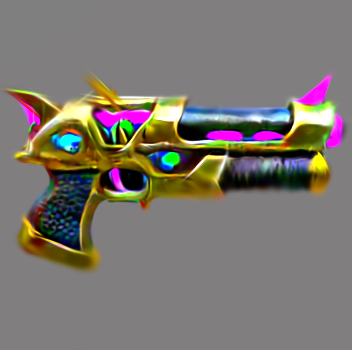}
						\put(65,67){{\includegraphics[width=0.04\linewidth]{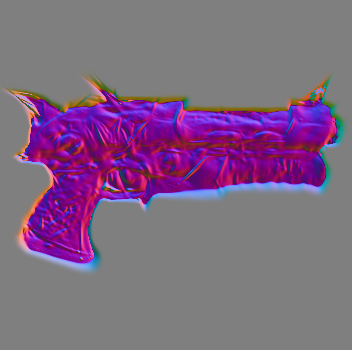}}}
					\end{overpic} &
        \begin{overpic}[width=0.13\linewidth]{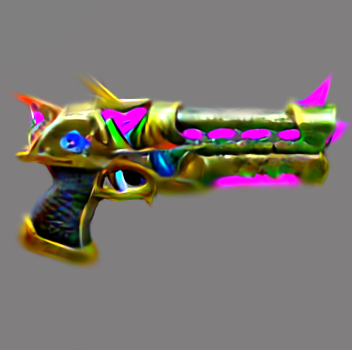}
						\put(65,67){{\includegraphics[width=0.04\linewidth]{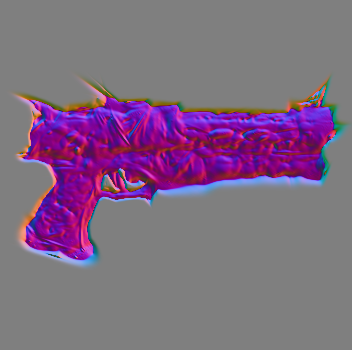}}}
					\end{overpic} &
        \begin{overpic}[width=0.13\linewidth]{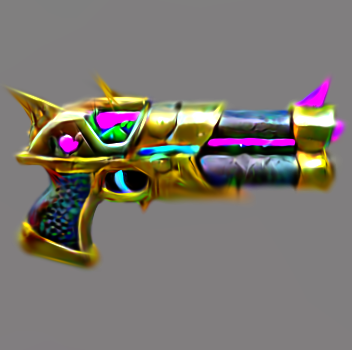}
						\put(65,67){{\includegraphics[width=0.04\linewidth]{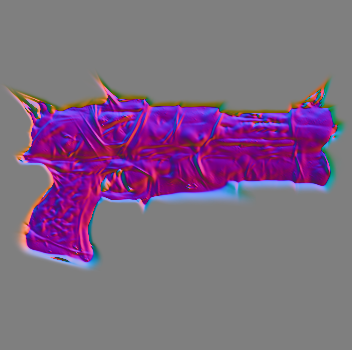}}}
					\end{overpic} &
        \begin{overpic}[width=0.13\linewidth]{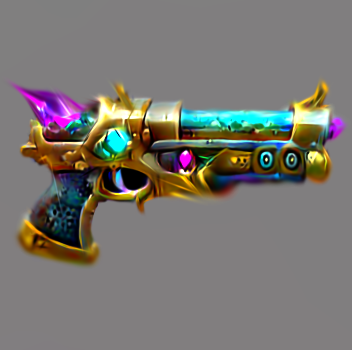}
						\put(65,67){{\includegraphics[width=0.04\linewidth]{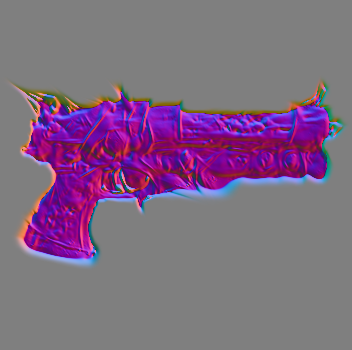}}}
					\end{overpic} &
        \begin{overpic}[width=0.13\linewidth]{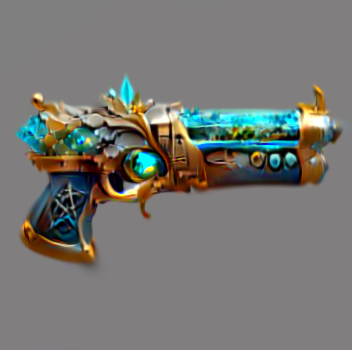}
						\put(65,67){{\includegraphics[width=0.04\linewidth]{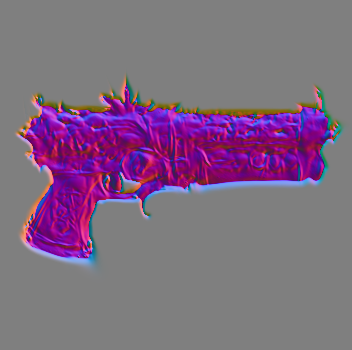}}}
					\end{overpic} &
        \begin{overpic}[width=0.13\linewidth]{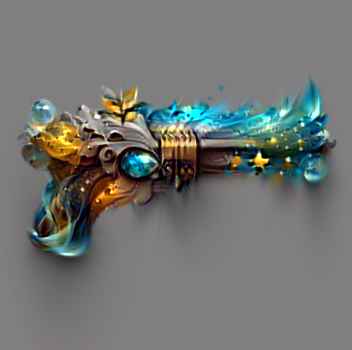}
						\put(65,67){{\includegraphics[width=0.04\linewidth]{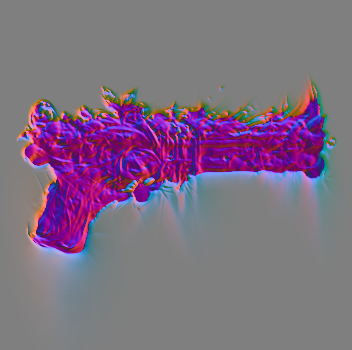}}}
					\end{overpic} &
        \begin{overpic}[width=0.13\linewidth]{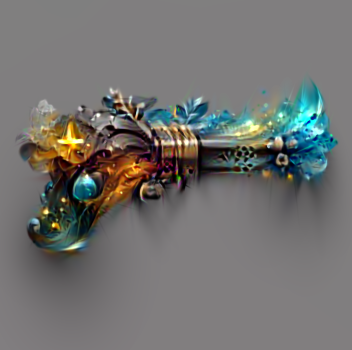}
						\put(65,67){{\includegraphics[width=0.04\linewidth]{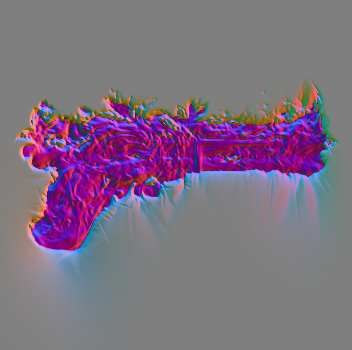}}}
					\end{overpic}
        \\
        \multicolumn{9}{c}{{(b) $\mathcal{L}= \alpha_1\mathcal{L}_{\mathrm{CDP}} + \alpha_3 \mathcal{L}_{\mathrm{ER}}$. The prompt is:  \prompts{Magic gun, game asset, mystery.}}}
    \end{tabular}
    \vspace{-0.2em}
    \end{tabular}}
        \label{subfig:diversity}
    \label{subfig:sds+er}
    \end{subfigure}
    \vspace{0.2em}

    \begin{subfigure}[t]{\textwidth}
        \centering
    \setlength{\tabcolsep}{1pt}
    \setlength{\fboxrule}{1pt}
    \resizebox{1\textwidth}{!}{
    \begin{tabular}{c}
    \begin{tabular}{ccccccccc}
        & $\alpha_3/\alpha_2=1e0$ & \multicolumn{6}{c}{\begin{tikzpicture}[baseline]  
                    \draw[->, >=latex, line width=0.35mm] (0,0.1) -- (14cm,0.1);  
                    \end{tikzpicture}  } & $\alpha_3/\alpha_2=1e4$
        \\
        \rotatebox{90}{\,\,\,\small{PickScore}} &
        \begin{overpic}[width=0.13\linewidth]{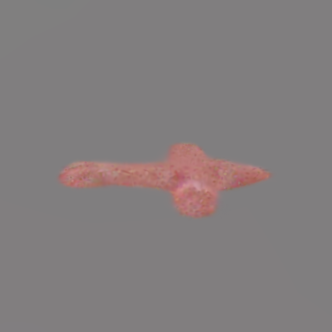}
					\end{overpic} &
        \begin{overpic}[width=0.13\linewidth]{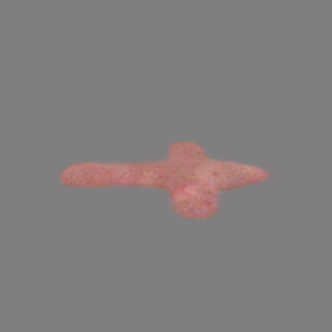}
					\end{overpic} &
        \begin{overpic}[width=0.13\linewidth]{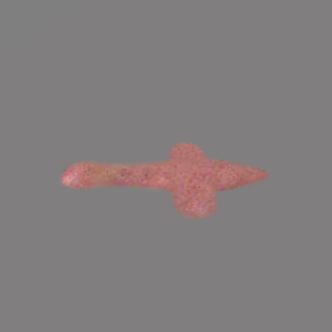}
					\end{overpic} &
        \begin{overpic}[width=0.13\linewidth]{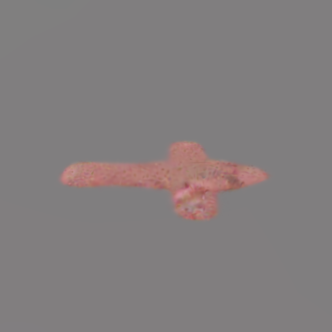}
					\end{overpic} &
        \begin{overpic}[width=0.13\linewidth]{./figure/cfg-2divergences/airplane-3_row1_col1.png}
					\end{overpic} &
        \begin{overpic}[width=0.13\linewidth]{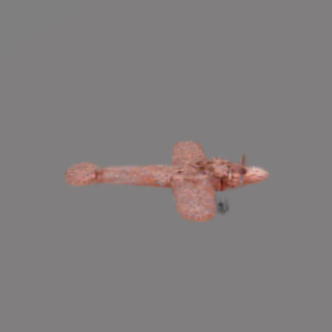}
					\end{overpic} &
        \begin{overpic}[width=0.13\linewidth]{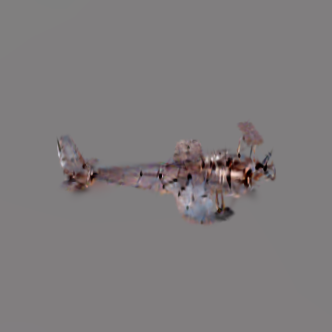}
					\end{overpic} &
        \begin{overpic}[width=0.13\linewidth]{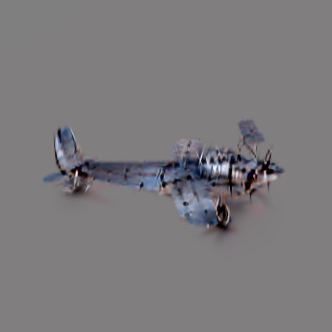}
					\end{overpic}
        \vspace{-0.2em}
        \\
        \multicolumn{9}{c}{{(c) $\mathcal{L}= -\alpha_2\mathcal{L}_{\mathrm{UDP}} + \alpha_3 \mathcal{L}_{\mathrm{ER}}$. The prompt is:  \prompts{An airplane made of metal.}}} \\
    \end{tabular}
    \vspace{-0.2em}
    \end{tabular}}
        \label{subfig:diversity}
    \label{subfigure:er-udp}
    \end{subfigure}
    \vspace{-0.5em}
    \caption{\textbf{Exploration of the design space of the unified divergence loss.} {We investigate their influence of different text-to-3D divergence losses on generation performance by adjusting their weights. 3D Gaussian splitting are used as representations in these experiments for computational efficiency.}}
    
    \label{fig:designspace}
    \vspace{-1em}
\end{figure*}

\paragraph{CDP and UDP}  
As shown in Fig.~\ref{fig:designspace}(a), assigning a large weight to the CDP loss preserves generation diversity but results in blurred outputs, while a large weight on the UDP loss introduces significant artifacts, preventing the 3D generation from capturing basic image features. Empirically, the optimal performance is achieved when the weights for CDP and UDP are balanced, i.e., when the ratio \(\alpha_1/\alpha_2\) is approximately one. This finding aligns with the observations for tuning the CFG loss\citep{wang2024prolificdreamer}, which typically attains the best performance when the key component, a balanced linear combination of CDP and UDP losses, is strongly emphasized.

\paragraph{CDP and ER}  
We employ two human-preference reward models, ImageReward~\citep{xu2023imagereward} and PickScore~\citep{kirstain2023pick}, in our experiments. As shown in Fig.~\ref{fig:designspace}(b), in both cases, higher ER scores lead to improved 3D generation by enhancing text alignment and revealing finer textual details. 
Different reward models can guide generation in distinct ways; for instance, in our test case, ImageReward and PickScore exhibit different preferences for color combinations that evoke a sense of mystery in the generated game assets.

\paragraph{UDP and ER}  
We further explore the combination of UDP and ER losses—a configuration not previously investigated in text-to-3D generation. As shown in Fig.~\ref{fig:designspace}(c), this novel combination also produces reasonable 3D objects, demonstrating that text-to-3D generation is possible without relying on a text-to-image model. However, we observe that its performance is highly sensitive to the initialization of the 3D representation; poor or biased initialization can trap the algorithm in local minima in the absence of text-to-image guidance. Consequently, although the ER+UDP combination is promising, integrating all three basic losses yields the best overall results.

\section{More Results}
\label{sec:app-moreresults}

\begin{figure*}[t] 
  \centering
  \resizebox{1.0\textwidth}{!}{ 
    \begin{subfigure}[t]{\textwidth}
      \centering
      \begin{tabular}{@{}cccc@{}}
        \begin{overpic}[width=0.24\linewidth]{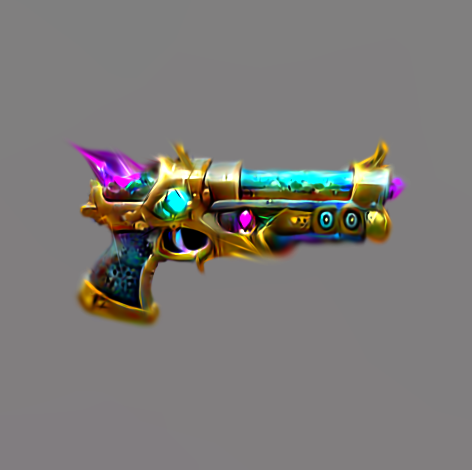}
          \put(65,65){{\includegraphics[width=0.08\linewidth]{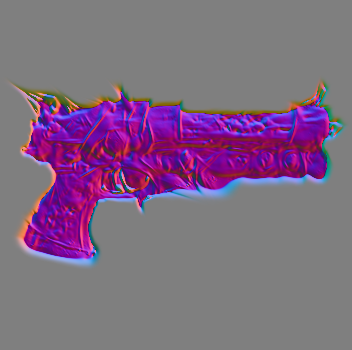}}}
        \end{overpic} &
        \begin{overpic}[width=0.24\linewidth]{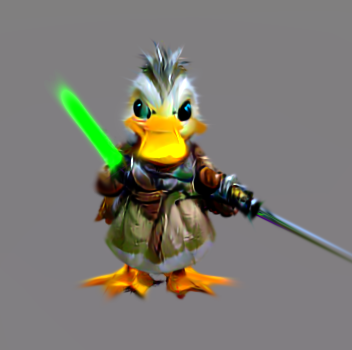}
          \put(65,65){{\includegraphics[width=0.08\linewidth]{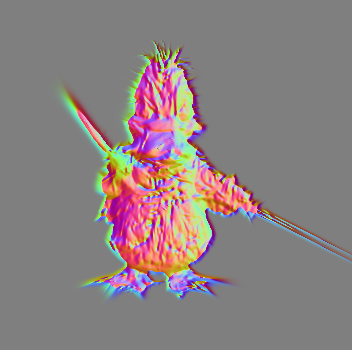}}}
        \end{overpic} &
        \begin{overpic}[width=0.24\linewidth]{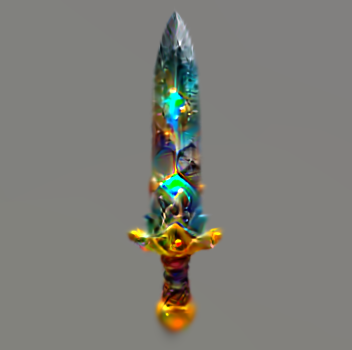}
          \put(65,65){{\includegraphics[width=0.08\linewidth]{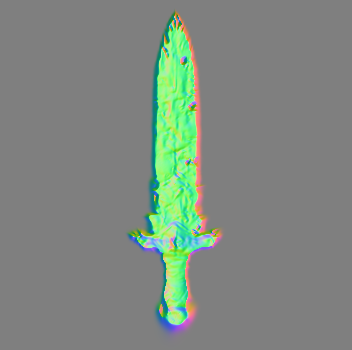}}}
        \end{overpic} &
        \begin{overpic}[width=0.24\linewidth]{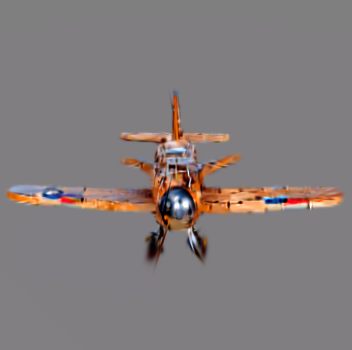}
          \put(65,65){{\includegraphics[width=0.08\linewidth]{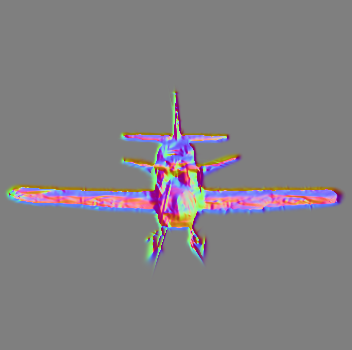}}}
        \end{overpic} \\
        \prompts{Magic gun, game asset, mystery.} &
        \prompts{Jedi duck holding a lightsaber.} &
        \prompts{Magic dagger, mystery, ancient.} &
        \prompts{An airplane made out of wood.} \\[5pt]
      \end{tabular}
      \label{subfig:gaussian_splattings_optimal}
    \end{subfigure}
  }
  \vspace{-0.5em}
  \caption{\textbf{Dive3D 3D Gaussian Splattings.} Dive3D can generate anime-style and game-style 3D Gaussian Splattings.}
  \label{fig:gaussian_splattings_optimal}
\end{figure*}

We provide more results in Figure~\ref{fig:comparisiongsalone}, \ref{fig:comparisiongs} and \ref{fig:gaussian_splattings_optimal}.

\vspace{-0.5em}

\end{document}